\documentclass[lettersize,journal,twoside]{IEEEtran}
\usepackage[numbers,sort&compress]{natbib}
\usepackage{graphicx} 
\usepackage{algorithmic}
\usepackage{amsfonts}
\usepackage{multirow}
\usepackage{comment}
\usepackage{amsfonts}
\usepackage{makecell}
\usepackage{amsmath}
\usepackage{graphics}
\usepackage{subfigure}
\usepackage{booktabs}
\usepackage{colortbl}  
\usepackage{xcolor}
\usepackage{array}  
\usepackage{array}  
\usepackage{hyperref}
\usepackage{verbatim}

\title{Exploring Gradient-Guided Masked Language Model to Detect Textual Adversarial Attacks}

\author{ Xiaomei Zhang, Zhaoxi Zhang, Yanjun Zhang, \\ Xufei Zheng, Leo Yu Zhang, Shengshan Hu, and Shirui Pan
\vspace{-0.2cm}
\thanks{

Correspondence to Dr. L. Zhang and Prof. X. Zheng 

Xiaomei Zhang, Leo Yu Zhang and Shirui Pan are with the School of Information and Communication Technology, Griffith University, Queensland, Australia (e-mail: xiaomei.zhang@griffithuni.edu.au, leo.zhang@griffith.edu.au, s.pan@griffith.edu.au).

Zhaoxi Zhang and Yanjun Zhang are with the School of Computer Science, University of Technology Sydney, Sydney, New South Wales, Australia (e-mail: Zhaoxi.Zhang-1@student.uts.edu.au, Yanjun.Zhang@uts.edu.au).

Xufei Zheng is with the College of Computer and Information Science, Southwest University, Chongqing, China (e-mail: zxufei@swu.edu.cn).

Shengshan Hu is with the School of Cyber Science and Engineering, Huazhong University of Science and Technology, Wuhan, China (e-mail: hushengshan@hust.edu.cn).
}}

\begin{document}

\date{June 2023}
\newcommand{\zzx}{\textcolor{blue}}

\maketitle

\begin{abstract}
Textual adversarial examples pose serious threats to the reliability of natural language processing systems. Recent studies suggest that adversarial examples tend to deviate from the underlying manifold of normal texts, whereas pre-trained masked language models can approximate the manifold of normal data. These findings inspire the exploration of masked language models for detecting textual adversarial attacks. We first introduce Masked Language Model-based Detection (MLMD), leveraging the mask and unmask operations of the masked language modeling (MLM) objective to induce the difference in manifold changes between normal and adversarial texts. Although MLMD achieves competitive detection performance, its exhaustive one-by-one masking strategy introduces significant computational overhead. Our posterior analysis reveals that a significant number of non-keywords in the input are not important for detection but consume resources. Building on this, we introduce Gradient-guided MLMD (GradMLMD), which leverages gradient information to identify and skip non-keywords during detection, significantly reducing resource consumption without compromising detection performance.
Extensive experiments show that GradMLMD maintains comparable or better performance than MLMD and outperforms existing detectors. 
Among defenses based on the off-manifold conjecture, GradMLMD presents a novel method for capturing manifold changes and provides a practical solution for real-world application challenges.

\end{abstract}

\section{Introduction}
\begin{figure*}[t]
    \centering
    \includegraphics[width=17cm]{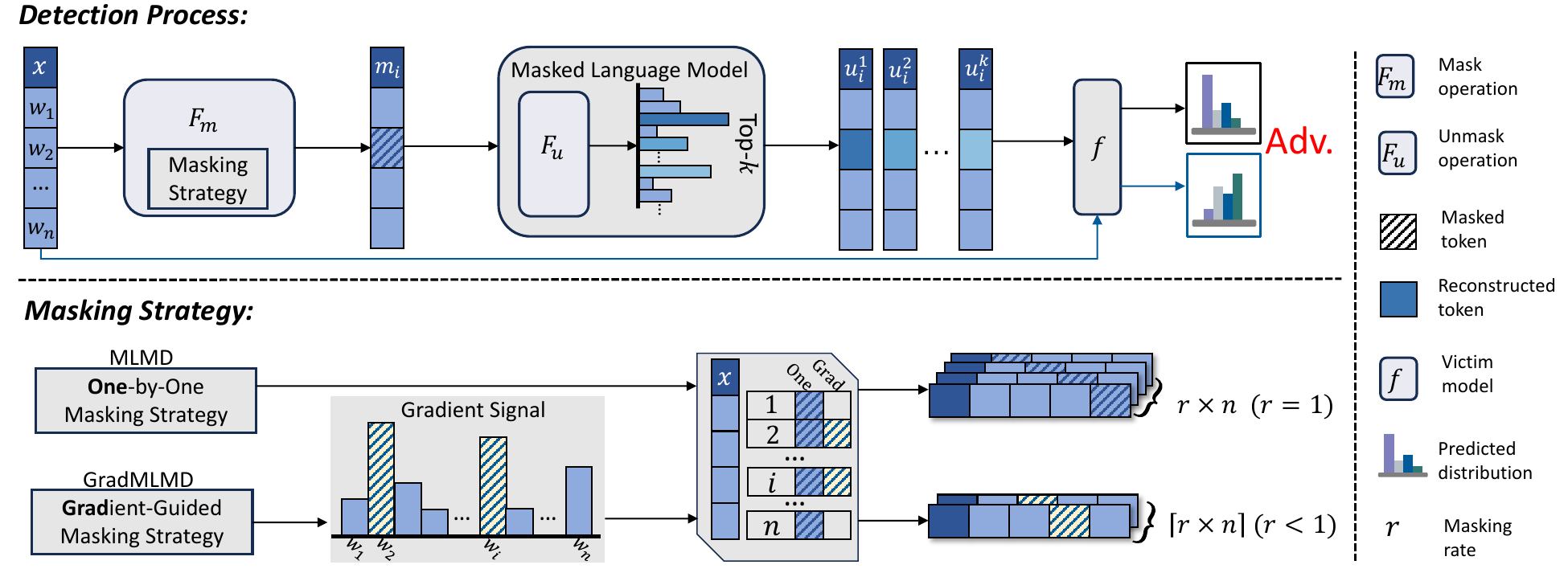}
    \centering
    \caption{An overview of Masked Language Model-based Detection (MLMD) and Gradient-guided MLMD (GradMLMD). 
    They share the same detection process, exploiting the differences in manifold changes between normal and adversarial examples induced by masked language models to detect adversarial attacks. 
    However, MLMD adopts a one-by-one masking strategy where every word in the input is masked individually, i.e., $r=1$.
    In contrast, GradMLMD employs gradient signals to assess the importance of each word. 
    It exclusively operates on keywords (yellow) in subsequent operations (i.e., $r<1$), thereby significantly reducing resource overhead.}
    
    \label{Fig:mlmdOverview}
\end{figure*}
\IEEEPARstart{A}{lthough} advanced deep neural networks have the potential to revolutionize the performance of myriad natural language processing (NLP) tasks~\cite{sun2025prompting, ding2025reshaping, gao2025perturbation}, they are highly vulnerable to adversarial attacks~\cite{Szegedy2014IntriguingPO, zhao2024disentangled,Zhang2022Evaluating,Zhang2024Stealing}. Through carefully manipulated inputs, attackers can drive models to produce erroneous outputs to their advantage. Many researchers have focused on introducing adversarial perturbations into the input by altering entire sentences. 
However, predominant efforts have been made to develop attacks at the word-level and character-level 
\cite{liu2022efficient,Ren2019GeneratingNL, Jin2020IsBR, Li2020BERTATTACKAA, Li2019TextBuggerGA, Gao2018BlackBoxGO, boucher2022bad}. The adversarial examples generated by these attacks either remain semantically invariant or closely resemble normal texts visually, making them difficult for humans to perceive.
%
Recent studies \cite{Gilmer2018AdversarialS,  Shamir2021TheDM, Tanay2016ABT, Zhang2023MaskedLM} have offered a new insight into understanding adversarial examples, suggesting that they tend to leave the underlying manifold of normal data. Consequently, the aforementioned adversarial behaviours can be interpreted as efforts to move normal examples off their original manifold by meticulously perturbing certain parts of the sentence.
The existence of adversarial examples presents substantial challenges to the integrity and reliability of NLP systems, emphasizing the urgent need for research on defense algorithms against such attacks.

To mitigate the vulnerability of NLP models (referred to as victim models in attack scenarios) to adversarial inputs, abundant defense techniques can be found in the NLP literature, including adversarial training \cite{Goodfellow2015ExplainingAH, li-etal-2023-text, ni2024dropattack}, input randomization \cite{Wang2022RethinkingTA,ebrahimi2025robustness}, synonym encoding \cite{Wang2021NaturalLA}, etc.
From the perspective of the data manifold, the effectiveness of these defenses stems from their ability to map adversarial examples as closely as possible to the manifold of normal data.
However, they require training from scratch 
or even modifying the model architecture. 

Instead of deploying robust new models, it would be more practical to distinguish between adversarial and normal examples.
Such strategy has several advantages over methods that
fortify the victim model's robustness.
Specifically, 
an adversarial detector can seamlessly integrate as a supplementary module without damaging the performance of the victim model. It identifies adversarial intentions, enabling appropriate responses of users. Furthermore, the detection algorithm can make defense methods more targeted by distinguishing which inputs are abnormal, thereby avoiding impact on normal inputs. It can also be effectively integrated with existing techniques \cite{Mozes2021FrequencyGuidedWS, Moon2022GradMaskGT}.
To detect adversarial texts, many detectors have made significant efforts in triggering changes in the manifold by substituting specific input tokens with synonyms or special tokens \cite{Mosca2022ThatIA, Moon2022GradMaskGT}. 
However, these examples with substituted tokens may not always be projected back onto the manifold where normal data clusters, 
making the observed manifold changes insufficient to construct effective detectors.

Recently, extensive research has indicated that pre-trained masked language models have the potential to capture information about the manifold of normal examples after completing a Masked Language Modeling (MLM) objective on abundant normal data 
\cite{Ng2020SSMBASM, Xu2021UnsupervisedOD}. The MLM task consists of two operations. The mask operation first stochastically perturbs the example off the current manifold by corrupting a percentage of tokens in the input. Then, an unmask operation is performed to project corrupted texts back on by training the model to restore the masked tokens using the remaining part of the sentence. These findings inspire us to explore the masked language model and the MLM objective to counter textual adversarial attacks.

In this work, we first present an initial method, Masked Language Model-based Detection (MLMD), to uncover the potential of pre-trained masked language models to detect adversarial texts by exploring changes in the manifold. 
Illustrated in Fig.~\ref{Fig:mlmdOverview}, MLMD employs a mask operation with a specific strategy to guide it away from its current manifold. Subsequently, the unmask operation projects the masked text back onto the normal data manifold. 
For an adversarial input, significant alterations occur in the input's manifold before and after these two operations. 
In contrast, for a normal data, as its manifold aligns well with the masked language model, performing detection operations does not induce any changes to its manifold.
MLMD stands out from other defense mechanisms that leverage changes in the manifold due to its utilization of masked language models. These models rely on the MLM objective to capture manifold features from an extensive corpus, thereby ensuring a superior approximation to the manifold of normal examples.

Although MLMD demonstrates better detection capability compared to the state-of-the-art defense mechanisms, it is computationally demanding. This is due to its failure to consider the difference in the contribution of words in the input to the detection results. It equally masks these words, employing a one-by-one masking strategy, leading to a significant number of ineffective mask and unmask operations. 

We thus undertake further investigations into MLMD to explore ways to enhance its practicality and facilitate easier deployment.  
We first conduct a posterior analysis for detection results in an ideal setup. This analysis affirms that a significant portion of input words are not important to the detection results of MLMD.  
Based on this, we categorize the words in the sentence into two groups: keywords, which are vital  for detection, and non-keywords, which are not essential for detection. Relying on an oracle masking strategy to eliminate these non-keywords from detection process, we can decrease resource overhead to a more practical level while preserving MLMD's  performance.

However, locating (non-)keywords during practical detection is challenging. 
Upon examining the implementation of adversarial attacks, we observe that the larger the gradient value, the more attention adversarial attacks will pay to these words. This marks them as pivotal indicators for adversarial detection \cite {Simonyan2013DeepIC, Moon2022GradMaskGT, Shen2023TextShieldBS}. 
Hence, we propose GradMLMD, which utilizes gradient information to identify non-keywords. 
As depicted in Fig.~\ref{Fig:mlmdOverview}, while GradMLMD follows the same detection framework as MLMD, it effectively excludes non-keywords in the mask operation, enabling it to detect adversarial inputs with just a few operations. 

Our experiments, carried out on three datasets with four victim models and against four representative attack methods, reveal that GradMLMD exhibits detection capabilities comparable to or even stronger than MLMD, outperforming the state-of-the-art competitors~\cite{Mozes2021FrequencyGuidedWS, Mosca2022ThatIA, Moon2022GradMaskGT} (Sec. \ref{subsec:detect_perf_mlmd}). Furthermore, we extend our experiments to explore the influence of masked language models, particularly the effects of the unmask operation, on detection. This investigation include various aspects, such as the impact of different masked language models (refer to Sec. \ref{subsec:various_models}), the removal or addition of the unmask operation (Sec. \ref{subsec:important_unmask}), the parameter settings of the unmask operation (Sec. \ref{subsec:setting_unmask}), and the fine-tuning of masked language models on specific downstream tasks to better align their manifolds (Sec. \ref{subsec:fine_tuning}).

The main contributions of this work can be summarized as follows:
\begin{itemize}
    \item We introduce MLMD, a method that detects malicious inputs by examining the changes in the manifold caused by mask and unmask operations in masked language modeling. It is the first work that reveals the capability of masked language models in detecting textual adversarial texts. 

    
    \item To improve the deployability and practicality of MLMD, we propose Gradient-guided MLMD (GradMLMD) to optimize the masking strategy. This method utilizes gradient information to identify and remove non-keywords, consequently reducing computation overhead.
    
    \item We conduct comprehensive experiments to assess the detection capabilities of MLMD and GradMLMD. Empirical results show that they achieve better performance compared to the state-of-the-art detection methods.
\end{itemize}

\section{Related Work}
\subsection{Textual Adversarial Attacks}

Due to text's discrete nature, textual adversarial perturbations often involve operations such as replacement, deletion, insertion, etc., on words, characters or sentences. 
Following previous works \cite{Ren2019GeneratingNL, Jin2020IsBR, Li2019TextBuggerGA}, we mainly focus on word-level and character-level attacks.
\subsubsection{Word-level Attacks} Word-level attacks aim to balance attack effectiveness, semantic coherence, and grammatical consistency . Approaches inspired by natural evolution employ population-based optimization techniques to identify appropriate perturbations. 
 Numerous studies \cite{Ren2019GeneratingNL, Jin2020IsBR} focus
on identifying critical words for substitution to enhance the stealth and effectiveness of attacks. Alternatively, some studies introduce models automatically generate perturbations to ensure linguistic fluency  \cite{ Li2020BERTATTACKAA}.


\subsubsection{Character-level Attacks} 
More fine-grained character-level attacks typically involve introducing typos in numbers, letters, and special symbols within the raw text. 
Despite semantic changes, resulting adversarial text visually resembles the original input and does not impact human judgment.
HotFlip \cite{Ebrahimi2018HotFlipWA} swaps one token for another by accessing the gradient. Meanwhile, TextBugger \cite{Li2019TextBuggerGA} and DeepWordBug \cite{Gao2018BlackBoxGO} first, identify critical parts within an input and disturb word characters appropriately.

From the perspective of manifold, these textual adversarial attacks elaborately engineer perturbations to steer examples away from the manifold of normal data. 
Sufficient perturbations, such as modifying enough characters or words, can cause the example to cross the decision boundary, resulting in incorrect predictions by the victim model.

\subsection{Defenses Against  Textual Adversarial Attacks}
Typical defenses for NLP victim models against adversarial attacks include enhancing robust predictions or detecting adversarial examples.

\subsubsection{Robust Prediction}
Adversarial training 
\cite{Goodfellow2015ExplainingAH,  li-etal-2023-text} augments normal examples with adversarial counterparts. However, this approach may not be feasible for deployed models.
Empirical studies \cite{Wang2022RethinkingTA,  gupta-etal-2023-dont} show that input randomization can neutralize most adversarial texts. The encoding method, instead, aims \cite{Wang2021NaturalLA} to ensure  similar encodings for similar inputs. However, these techniques might compromise model performance on clean datasets.
Recent defenses focusing on manifold assumptions have gained attention \cite{Meng2017MagNetAT, Zhang2021SelfSupervisedAE}. TMD \cite{Nguyen2022TextualMD}, for instance, employs a generative model to project all inputs onto the learned normal manifold.

Considering the manifold, adversarial training aims to regulate examples near the manifold boundary. Randomization-based and encoding-based methods endeavour to map malicious examples back onto the normal manifold, thereby mitigating their adversarial effects.

\subsubsection{Adversarial Detection}
Adversarial detection focuses on identifying potential threats within the input. 
Character-level attacks are countered by spell-checking 
systems \cite{Pruthi2019CombatingAM} designed to detect and correct erroneous characters.
For word-level attacks, DISP \cite{Zhou2019LearningTD} utilizes a discriminator and embedding estimator to detect and correct adversarial parts.
FGWS \cite{Mozes2021FrequencyGuidedWS} is constructed based on the assumption that adversarial attacks prefer exploiting infrequent words exposed in the training set. 
Several recent studies, including WDR \cite{Mosca2022ThatIA} and GRADMASK \cite{Moon2022GradMaskGT}, detect adversarial behaviors by analyzing how the victim model's responses change when input tokens are substituted with synonyms or special tokens.

From the manifold perspective, a detector only needs to distinguish between off-manifold and on-manifold (i.e., normal) examples. Detectors for both character-level attacks and word-level attacks can be seen as mechanisms that promote manifold changes, crucial for shaping effective detectors. 
Token replacement and spell-checking operations do not ensure that the input is guided back to the manifold of normal data after detection operations.
Therefore, a projection method that accurately captures the normal data manifold will naturally improve the features to develop effective adversarial detectors.

\subsection{Masked Language Models and the Off-manifold Conjecture}
\label{Sec:Conjecture}

Masked language models have spurred significant advancements in understanding natural language. 
Pre-trained on a substantial amount of unlabeled normal data using the Masked Language Modeling (MLM) objective, these models acquire the capacity to reconstruct input text.
BERT \cite{Devlin2019BERTPO} is the first bidirectional language model trained via the MLM task, designed to learn universal language representations.
ALBERT, a lightweight version of BERT \cite{Lan2020ALBERTAL}, employs n-gram masking in its MLM objective, covering entire words up to n-grams. This approach enhances its ability to grasp language comprehensively despite sharing pre-training data with BERT.
RoBERTa \cite{Liu2019RoBERTaAR} improves upon BERT with a dynamic masking strategy that adapts masking patterns dynamically during training. It also undergoes pre-training on diverse corpora, utilizing advanced techniques like extended training durations and larger batch sizes.
Thus, it can fit the normal manifold more accurately.

The off-manifold conjecture \cite{Shamir2021TheDM, Gilmer2018AdversarialS, Tanay2016ABT} indicates that adversarial examples do not lie inside the data manifold of normal examples.
This conjecture offers an alternative perspective for interpreting the existence of adversarial examples and has garnered considerable research interest. 
In computer vision tasks, numerous studies 
\cite{Meng2017MagNetAT, Zhang2021SelfSupervisedAE} validate this conjecture and develop corresponding defensive approaches.
In NLP, the defense method \cite{Nguyen2022TextualMD} confirms the validity of the conjecture in the contextualized embedding space of textual examples.

Concurrently, many studies \cite{ Ng2020SSMBASM, Xu2021UnsupervisedOD} show that masked language models can effectively model the manifold of normal textual data and can be leveraged to improve out-of-distribution robustness. 
These findings motivate us to explore a novel approach to detect textual adversarial attacks. 
In particular, by harnessing the capability of masked language models to fit the normal data manifold, we hypothesize that the behaviour of off-manifold (i.e., adversarial) examples will exhibit differences from normal ones when processed by masked language models.

\section{Method}
\subsection{Notations}
\label{subsec:Notations}
In this work, we aim to detect adversarial examples generated to deceive standard classification models. The defender is assumed to have full access to the victim model, such as parameters and architecture. Moreover, if a given input is adversarial, the defender is not required to know which specific attack method was used to generate it. 
The victim model $f(\cdot)$ is trained on the input sequence $x$ and its ground-truth label 
$ y^* \in \{1, 2, \cdots, c\}$, where
$c$ is the number of classes. 
In the inference phase, $f$ outputs the confidence score vector $f(x)$, where $\sum_{y=1}^c f(x)_y = 1$. 
The final prediction result is $z(x)=\underset{y}{\arg\max}\, f(x)_y$.

A normal input $\tilde{x}$ and its corresponding adversarial counterpart $\hat{x}$ should satisfy
$\hat{x} = \tilde{x}+\delta$, $z(\hat{x})\neq z(\tilde{x})$.
Here, $\delta$ is the adversarial perturbation. 
In NLP, meaningful and imperceptible perturbation is typically achieved by adding, removing, or substituting words or characters in the raw normal input $\tilde{x}$. Significantly, $\delta$ is essentially generated by an iterative optimization, which will be detailed in Sec.~\ref{subsec:gradient}. 

\subsection{Masked Language Model Based Detection - MLMD}
\label{Sec:mlmd}

In this section, we present our initial method, MLMD, to validate the effectiveness of using the masked language model for adversarial input detection. 
MLMD involves three parts: 
\begin{itemize}
    \item $F_m$: A mask function corrupts raw input based on the masking strategy, moving the text away from its original manifold;
    \item $F_u$: An unmask function leverages the masked language model $\Phi$ to map the masked texts produced by $F_m$ back to the manifold of normal examples;
    \item $C_a$: An adversarial classifier that detects adversarial examples by capturing the differences in manifold changes.
\end{itemize}
Formally, our detector is a distinguisher $\mathrm{d}(F_m, F_u, C_a)$: $\tilde{\mathcal{X}}\cup \hat{\mathcal{X}}\rightarrow\mathcal{Y}$, where $\tilde{\mathcal{X}}$ and $\hat{\mathcal{X}}$ are the space of all normal texts and adversarial texts, respectively.
$\mathcal{Y}=\{0, 1\}$ is the set of ground truth binary labels, with $1$ indicating adversarial examples. Given the intractability of capturing the entire space of normal and adversarial data, we estimate $\tilde{\mathcal{X}}\cup\hat{\mathcal{X}}$ by leveraging the dataset
$\mathcal{D}=\tilde{\mathcal{D}} \cup \hat{\mathcal{D}} $, 
where $\tilde{\mathcal{D}}$ is composed of normal data $\tilde{x}$, and $\hat{\mathcal{D}}$ consists of adversarial counterparts $\hat{x}$.

\subsubsection{Mask and Unmask Operations}
\label{subsubSec:mask_and_unmask}
For any input $x\in \mathcal{D}$, composed of $n$ words $\{w_{\textnormal{1}}, w_{\textnormal{2}}, \cdots, w_{n}\}$, $F_m$ maps it into the masked manifold. This process generates an ensemble of masked sequences, which can be expressed as:
\begin{equation}
    M=F_m(x,r).  
\label{Eq:mask}
\end{equation} 

The mask function $F_m$ outputs $M=\{m_i|i\in  [1, \lceil r \times n \rceil] \} $, where each $m_i$ represents i-th masked text generated by replacing the selected token in $x$ with $[MASK]$, with $r\in (0,1]$ denoting the masking rate and $n$ representing the length of $x$.

We note that the implementation of $F_m$ is not restricted; it can either individually mask words of $x$ (with $r=1$) one by one, a strategy employed in instantiating our MLMD, or involve more sophisticated masking strategies which will be elaborated in detail in Section~\ref{Sec:Gradmlmd}. 

Subsequently, a masked language model $\Phi$ is employed to reconstruct the corrupted words for texts in $M$.
For each $m_i$, we preserve the top-$k$ candidates recovered by $\Phi$:
\begin{equation}
    U=F_u(M,\Phi, k ),   
    \label{Eq:unmask}
\end{equation}
where $U=\{u_i^j| i\in [1, \lceil r \times n \rceil], j\in [1,k]\}$ indicates the set of reconstructed sequences. 
Each $u_i^j$ is the $j$-th ($j \in [1, k]$) rebuild candidate when unmasking the text $m_i$. 

Recall the discussions in the Sec.~\ref{Sec:Conjecture}, it becomes evident that for a normal example, the manifold remains unchanged through $F_m$ and $F_u$. This contrasts with the change from an off-manifold to an on-manifold state when $x$ is adversarial.

\subsubsection{Building Threshold-based Classifier}
The mask and unmask operations produce significantly distinguishable signals for normal and adversarial examples. 
We will now investigate the utilization of these signals in the detection of adversarial attacks. 
A distinguishable score $S(x, f, \Phi)$ is defined for input $x$ based on the masked language model $\Phi$ and the victim model $f$:
\begin{equation}
    S(x, f, \Phi)= \frac{1} {  n\times k  }  \sum\limits_{i=1}^{\lceil r\times n \rceil}\sum\limits_{j=1}^{k}\mathbb{I}(z(x), z(u_i^j)),
    \label{Eq:disScores}
\end{equation}
where $z(x)=\underset{y}{\arg\max}\, f(x)_y$, $u_i^j \in U$  (defined by Eq.~(\ref{Eq:unmask})),
and
{$\mathbb{I}(\cdot, \cdot)$}
is the indicator function, which yields $0$ when the two operands are equal. 

Clearly, $S(x, f, \Phi)$ falls into $[0, 1]$.
According to the manifold conjecture outlined in Sec.~\ref{Sec:Conjecture}, $S(x, f, \Phi)$ tends to be small when $x$ is on-manifold (i.e., normal examples) and larger when $x$ is off-manifold. Visual validation of this concept can be observed in Fig.~\ref{Fig:D_scores}.
After computing the score $S(x, f, \Phi)$ for the dataset $\mathcal{D}$, obtaining the desired adversarial classifier $C_a$ is straightforward; it involves simply selecting a suitable threshold $\tau$:
\begin{equation}
\mathtt{C_a}(x)=\begin{cases}
     $0$ \quad \mathrm{if} \quad  S <=\tau \\
     $1$ \quad \mathrm{else}
\end{cases}
\end{equation}
where $\tau$ is empirically determined through a \textit{one-time} offline process by electing the value that maximizes the F1 score. This threshold is subsequently employed for online detection.
\subsubsection{Building Model-based Classifier}
\label{subsubsec:ModelbasedDetector}
The threshold-based classifier captures manifold information at the label level. Similar to prior studies \cite{ Carlini2021MembershipIA}, we also propose a method that leverages confidence scores from the victim model to capture manifold changes.
It consists of two steps in the following. 

\textbf{Feature Engineering.} For an input $x$, there will be $k \times \lceil r \times n \rceil$ reconstructed candidates after $F_m$ and $F_u$. Inspired by the work \cite{Mosca2022ThatIA}, a feature vector $FE = [fe_l]_{l=1}^{k \times \lceil r \times n \rceil}$ for $x$ can be obtained:
\begin{equation}
    fe_l= f(u_i^j)_{y^*} - \underset{y \neq y^*}{\max}f(u_i^j)_y, 
\label{Eq:featureVector}
\end{equation}
where $y^* = \underset{y}{\arg\max}\, f(x)_y$, $i \in [1, \lceil r \times n \rceil]$, $j \in [1, k]$, and $l = (i-1)\times k + j$.
It's evident that if $x$ is normal, $fe_l$ tends to be positive; conversely, if $x$ is adversarial, $fe_l$ tends to be negative.

\textbf{Binary Classifiers.} With the features available, we  develop the dataset as $\Gamma = \{ (FE_{\tilde{x}}, 0)\} \cup \{(FE_{\hat{x}}, 1)\} $ for $\tilde{x} \in \tilde{\mathcal{D}}$ and $\hat{x} \in\hat{\mathcal{D}}$, and train a binary classifier based on it.  
To investigate how the feature vector's element order impacts the detection score, we  sort the feature vector $FE$ in ascending order, denoting  the sorted feature vector as $\overline{FE}$. We then construct $\overline{\Gamma} = \{ (\overline{FE}_{\tilde{x}}, 0)\} \cup \{(\overline{FE}_{\hat{x}}, 1)\}$ and also train a binary classifier based on this sorted dataset. 
The performance comparison of the original and sorted feature vectors is detailed in Sec.~\ref{Sec:model_based_performance}. 
In addition, to ensure training consistency regarding input dimensions, 
we either pad the example with ones or truncate it to the appropriate length.

\subsubsection{Limitation of MLMD} 
\label{subsubsection:limitation}
In MLMD, words in the input are treated equally, employing a one-by-one masking strategy in mask operation $F_m$ (i.e., $r=1$). It generates $n$ masked sequences for $n$ words and reconstructs each word $k$ times, leading to $n$ interactions with the masked language model and $ n \times k $ interactions with the victim model. While these operations ensure MLMD's proficiency in detecting adversarial texts, they raise concerns about resource costs.


\subsection{GradMLMD}
\label{Sec:Gradmlmd}

\begin{figure}[t]
    \centering 
        \subfigure[]{%
        \includegraphics[width=4.30cm]{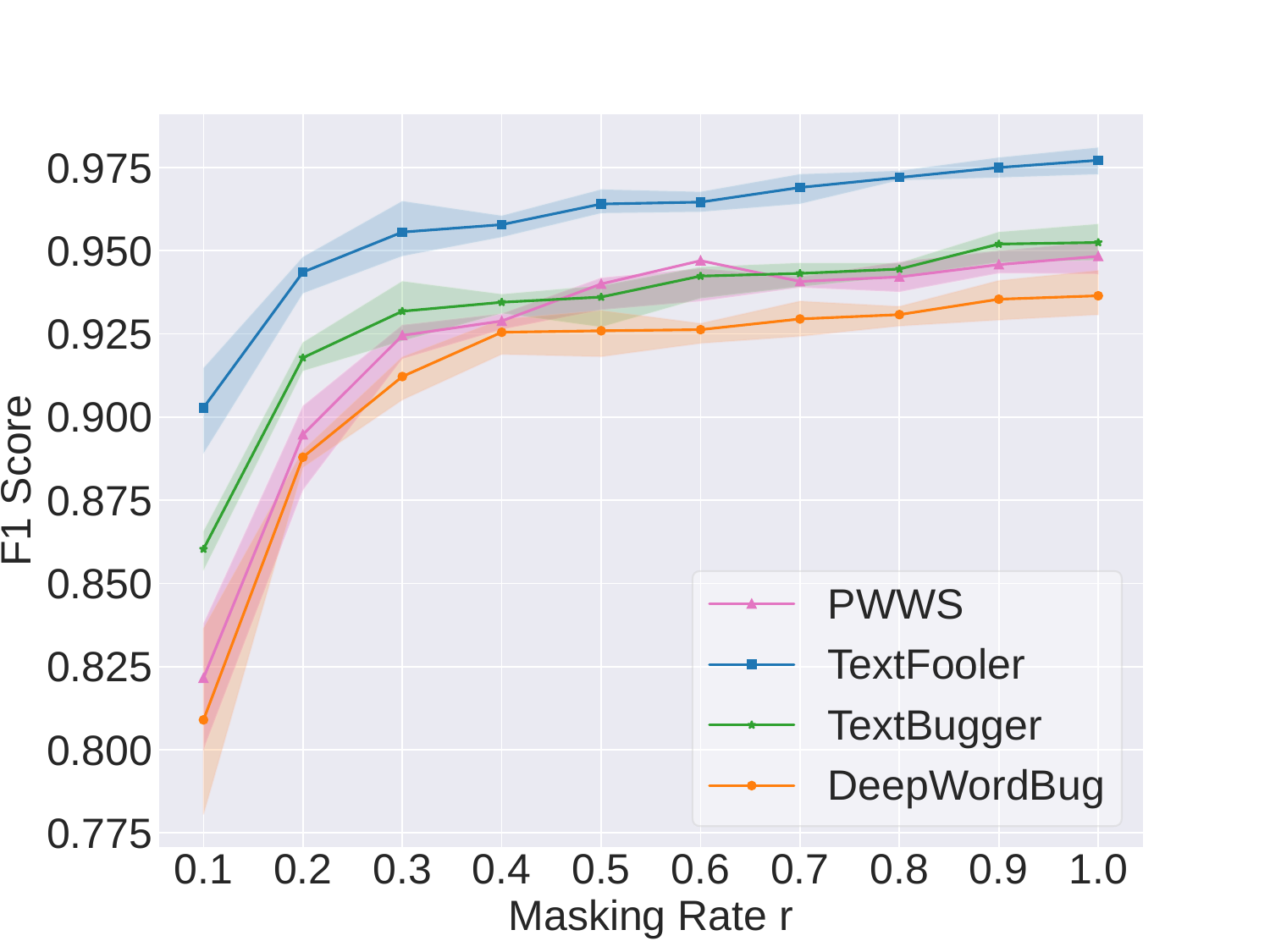}
        \label{Fig:masking_rate_f1}}
        \subfigure[]{%
        \includegraphics[width=4.2cm]{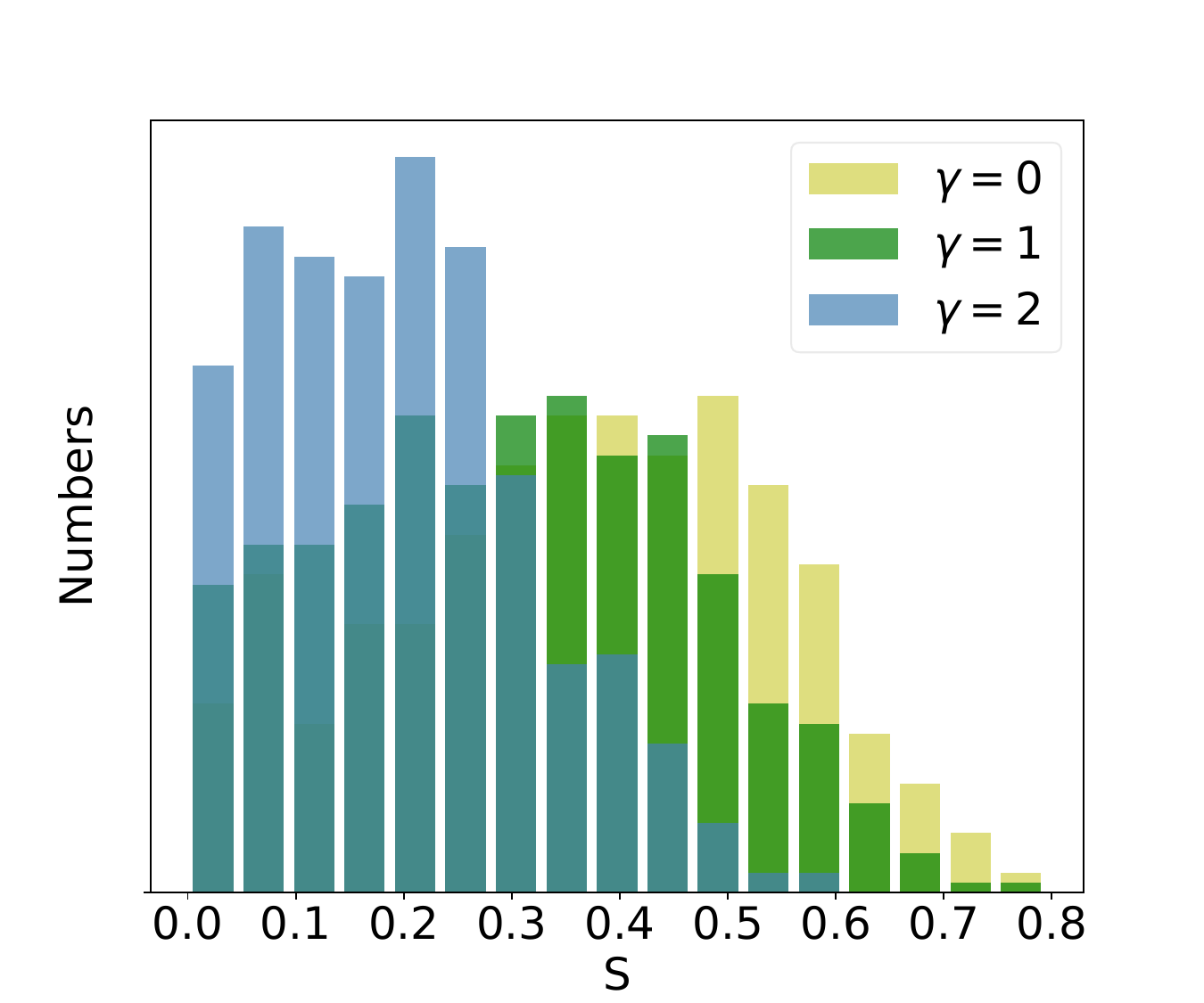}
        \label{Fig:adv_dis_score}}
       \caption{(a) The effect of the masking rate of the one-by-one masking strategy on detection performance. Results averaged over {$5$} runs. (b) The distribution of distinguishable score $S$ for adversarial examples with different $\gamma$. This experiment is carried out with the AG-NEWS-TextFooler-BERT configuration. The y-axis represents the number of adversarial examples. 
       }
\end{figure}

To enhance the practicality of MLMD, we systematically examine the steps of the detection process and perform experiments using fine-tuned BERT on AG-NEWS with four attack methods. We investigate a variety of masking rates, ranging from $0.1$ to $1.0$, aiming to observe the impact of words on the detection outcomes.


Illustrated in Fig.~\ref{Fig:masking_rate_f1}, it's evident that masking rates ranging from $0$ to $0.5$ result in significant enhancements in F1 score. However, beyond $0.5$, the improvement in detection performance becomes less noticeable. This observation implies that some words are pivotal for detection, while the rest are deemed less important (i.e., non-keywords). 
If we can filter out these non-keywords, computational resources can be conserved without sacrificing detection ability.


\subsubsection{Oracle Method}
\label{subsubsec: oracle_method}

We now investigate the existence of non-keywords and explore their impact on detection results in MLMD.
After knowing predictions (from the victim model) of $x$ and its $n \times k$ unmasked texts, a posterior analysis is conducted based on the observed detection results. 
In this setting, 
we select certain words from input $x$ to create the oracle non-keyword set $O$, ensuring these words satisfy a specific criterion: 
\begin{equation}
    \sum\limits_{j=1}^{k}\mathbb{I}(z(x), z(u_i^j))<=\gamma,
    \label{Eq:orcal_gemma}
\end{equation} 

where $\mathbb{I}(\cdot, \cdot)$ is the indicator function,  which yields $0$ when the two operands are
equal. Here, $u_i^j$ means the j-th rebuild text for masked example $m_i$. 
After unmasking, these words do not change the victim model's predictions, consequently leaving the distinguishable score of the entire sentence $x$ unaffected. 
Therefore, they are considered unhelpful in identifying adversarial examples.
In our analysis, 
we set $k=3$, as this setting achieves the optimal detection performance (detailed in 
Sec. \ref{subsec:setting_unmask}). Consequently, $\gamma \in \{0,1,2,3\}$. 


We aim to select a $\gamma$ that does not significantly affect the detection ability while effectively filtering as many non-keywords as possible.
Initially, we assess the influence of $\gamma$ on detection results.
The distribution of distinguishable scores $S$ between normal and adversarial examples serves as an intuitive reflection of detection performance. 
Consequently, this part assesses how $\gamma$ impacts detection ability via the score distributions.
Importantly, the score for input $x$ is determined by the remaining words in the input after removing non-keywords within the set $O$ formed by a specific $\gamma$. 
Due to the capability of the masked language model to accurately capture the normal manifold, minimal changes occur in the manifold of normal inputs after mapping. As a result, their scores predominantly cluster around $0$ across all $\gamma$ settings. Therefore, we focus on examining the distribution of $S$ for adversarial examples to reflect disparities in detection ability. 
If these distributions tend toward $0$, the overlap between the distributions of normal and adversarial inputs will increase, making it harder to distinguish between them.

Fig.~\ref{Fig:adv_dis_score} illustrates the distribution of distinguishable scores for adversarial examples across various values of  $\gamma$ in the configuration ``AG-NEWS-TextFooler-BERT" (i,e., denoting adversarial examples crafted by attacking BERT fine-tuned on AG-NEWS using TextFooler.). 
When $\gamma=0$,  
eliminating non-keywords does not impact detection ability;  
the distribution of these scores is equivalent to the distribution of adversarial examples obtained in MLMD.
Similarly, for $\gamma=1$, detection ability remains unaffected due to its significant overlap with $\gamma=0$ in most regions. 
In contrast, 
at $\gamma=2$, distinguishable scores primarily cluster in a lower-value range, markedly unfavorable for detection purposes.
We omit $\gamma=3$ since, under this condition, all words are treated as non-keywords and removed. 
Therefore, when evaluating the influence on detection capability, $\gamma=0$ and $\gamma=1$ stand out as feasible choices.

We then investigate the proportion of non-keywords for various $\gamma$. 
We use $\tilde{pr}=\frac{1} {|\tilde{\mathcal{D}}|}  \sum\limits_{i=1}^{|\tilde{\mathcal{D}}|}\dfrac{|\tilde{O}_i|}{|\tilde{x}_i|}$ and $\hat{pr}=\frac{1} {|\hat{\mathcal{D}}|} \sum\limits_{i=1}^{|\hat{\mathcal{D}}|}\dfrac{|\hat{O}_i|}{|\hat{x}_i|}$ to
represent the non-keyword proportions for normal and adversarial texts, respectively, where $\tilde{O}_i$ and $\hat{O}_i$ are the non-keyword sets formed by a specific $\gamma$ for a normal text $\tilde{x}_i$ (from $\tilde{\mathcal{D}}$) and its adversarial text $\hat{x}_i$ (from $\hat{\mathcal{D}}$), and $|\cdot|$ is used to calculate the length. 
We note that normal examples and adversarial examples appear in pairs, which means $|\tilde{\mathcal{D}}|=|\hat{\mathcal{D}}|$ in dataset $\mathcal{D}$. 
The final proportion of non-keywords  is determined by $\min(\tilde{pr},\hat{pr})$.

\begin{table}[htbp]
  \centering
  \caption{The proportion of non-keywords in oracle non-keyword set under different $\gamma$ settings to the input. }
    \begin{tabular}{cccc}
    \toprule
    $\gamma=0$ & $\gamma=1$ & $\gamma=2$ & $\gamma=3$ \\
    \midrule
    0.458 & 0.699 & 0.854 & 1.000 \\
    \bottomrule
    \end{tabular}%
  \label{tab:overlap_gamma}%
\end{table}%

Table~\ref{tab:overlap_gamma} shows the results under the AG-NEWS-TextFooler-BERT configuration, which highlights that at $\gamma=0$, the proportion of non-keywords is $46\%$, whereas it increases to $70\%$ at $\gamma=1$. Taking into account the previous findings on the influence of $\gamma$ on detection capability, $\gamma=1$ is selected as it retains effective detection performance while maximizing the number of non-keywords that can be removed from the detection process.
We named the technique of selecting non-keywords when $\gamma=1$ the oracle method. 
The strategy of filtering out the non-keywords with this method in the mask operation is referred to as the oracle masking strategy.


\begin{figure}[t]
    \centering 
         \subfigure[]{\label{Fig:overlap_nonkey_bert_SUM1}
        \includegraphics[width=4.25cm]{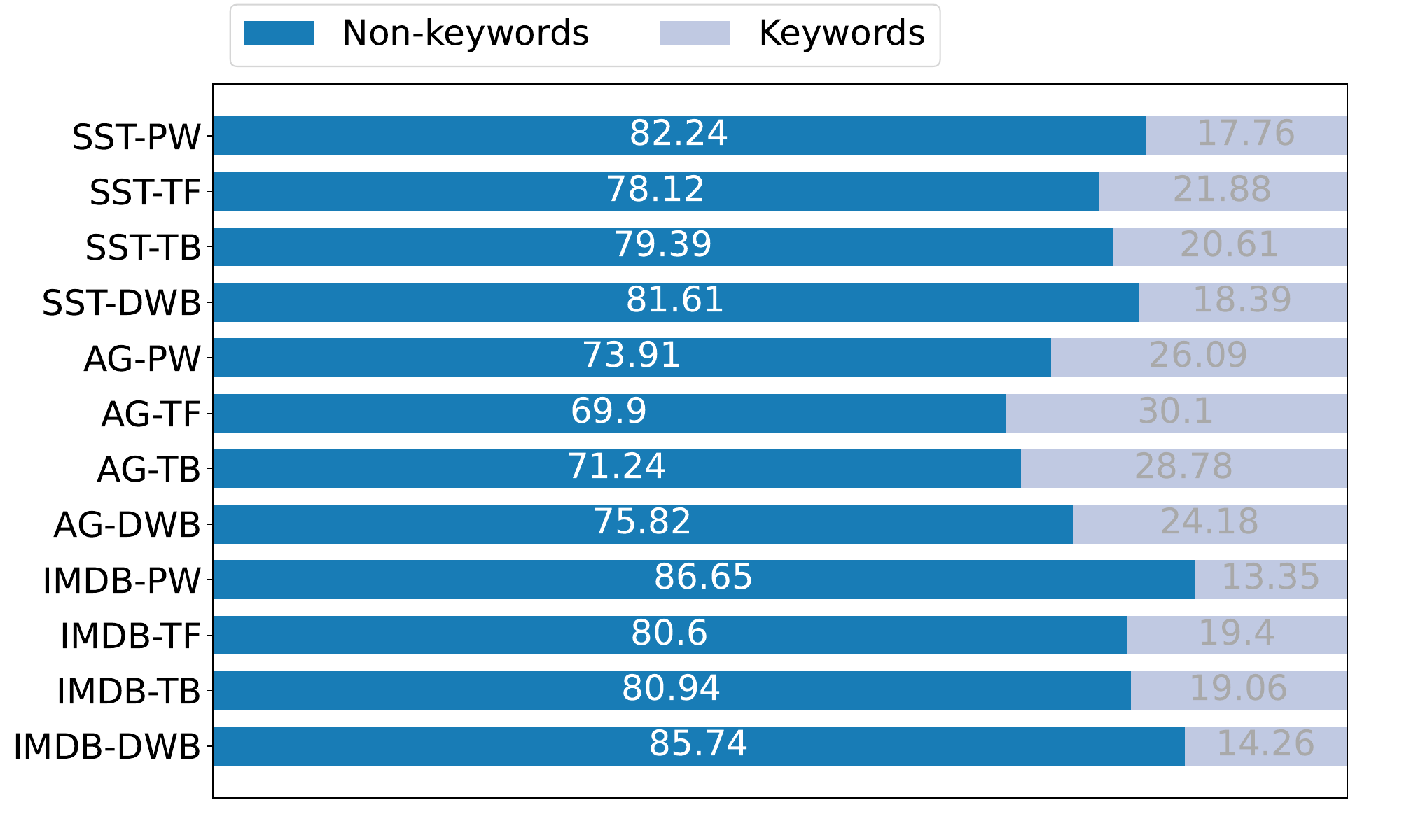}}
        \subfigure[]{%
        \label{Fig:grad_orcle_nonkey_bert_SUM1}
        \includegraphics[width=4.25cm]{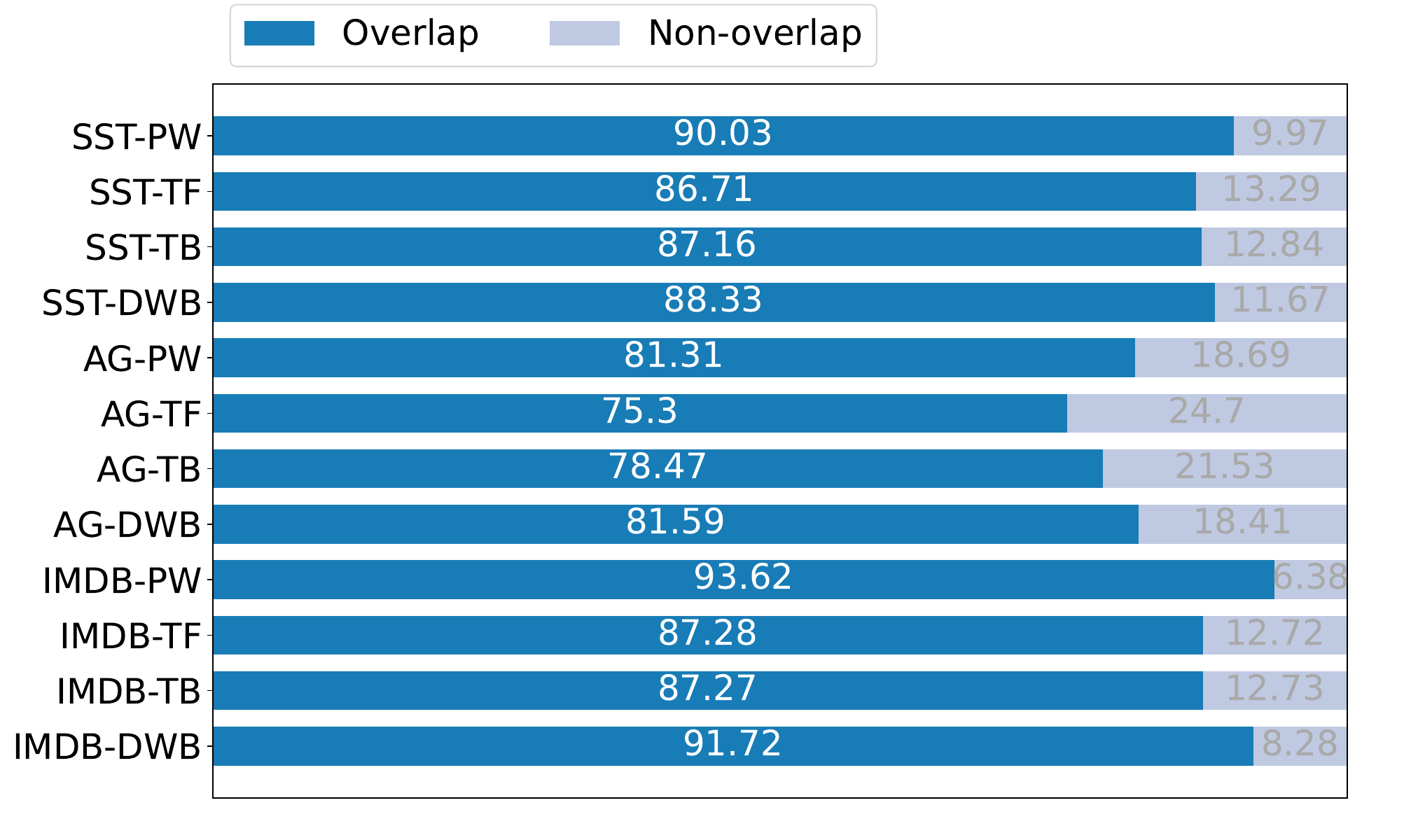}}
       \caption{(a) The proportion of non-keywords selected by the oracle method to the input. (b) The overlap rate between non-keywords is identified by the oracle method and the gradient-guided method. The victim models for both experiments are fine-tuned BERT. We use SST, AG to represent SST-2, AG-NEWS, respectively, and PW, TF, TB, and DWB stand for PWWS, TextFooler, TextBugger and DeepWordBug, respectively.
       } 
\end{figure} 

We also show the proportion of non-keywords across various configurations using the oracle method. The results in Fig.~\ref{Fig:overlap_nonkey_bert_SUM1} highlight that a minimum of $70\%$ of the words within the input are not crucial for detection. 
This insight serves as a reference for empirically determining the number of words considered non-keywords in Sec.~\ref{subsec:detect_perf_gradmlmd}. 
Additionally, the proportion in the AG-NEWS is generally lower on average compared to SST-2 and IMDB. This difference could be attributed to its nature as a four-class classification task, which is inherently more complex than a binary classification task and thus requires more keywords.

\begin{figure}[t]
    \centering 
        \subfigure[SST-2]{\includegraphics[width=4.35cm]{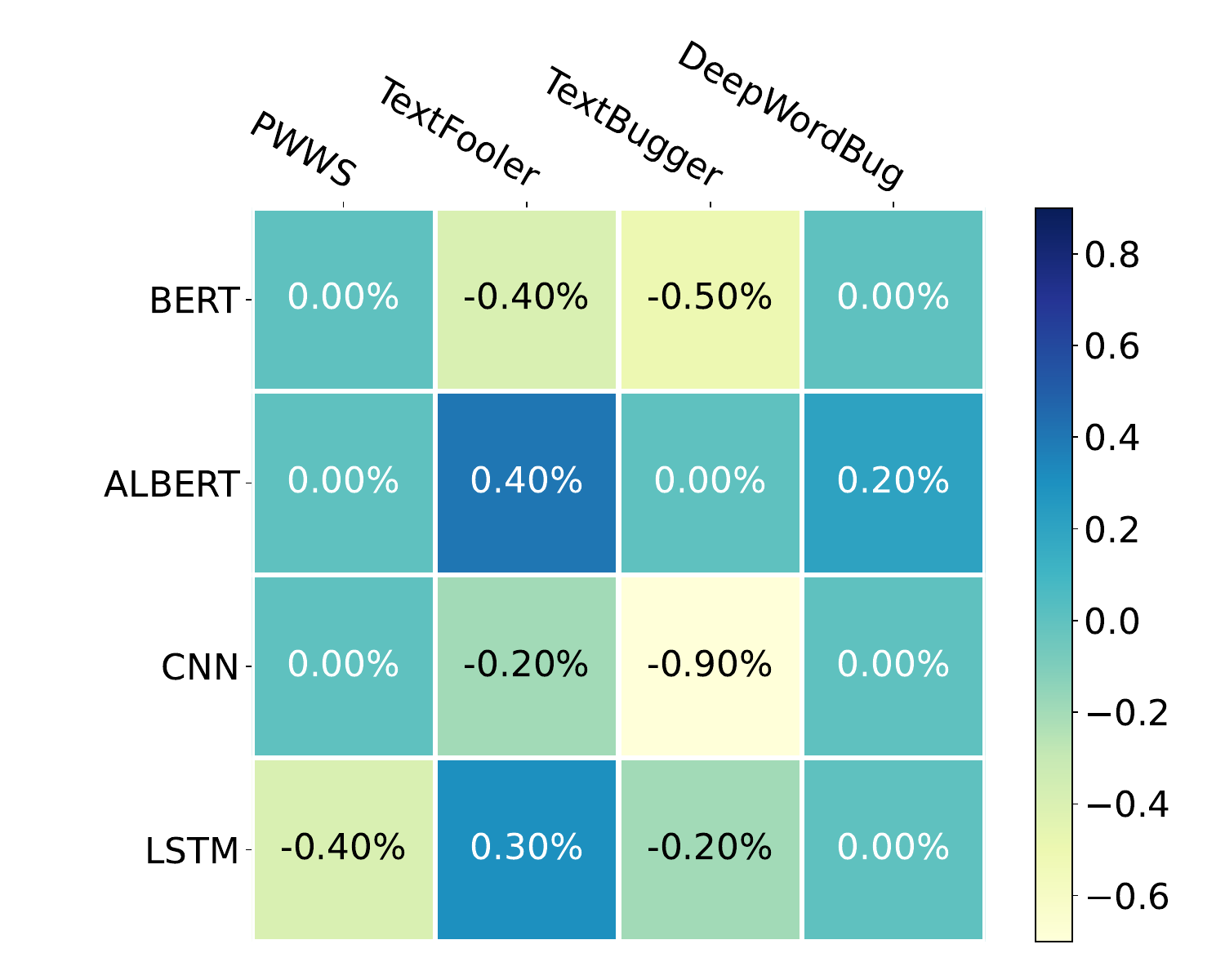}}
        \subfigure[AG-NEWS]{\includegraphics[width=4.35cm]{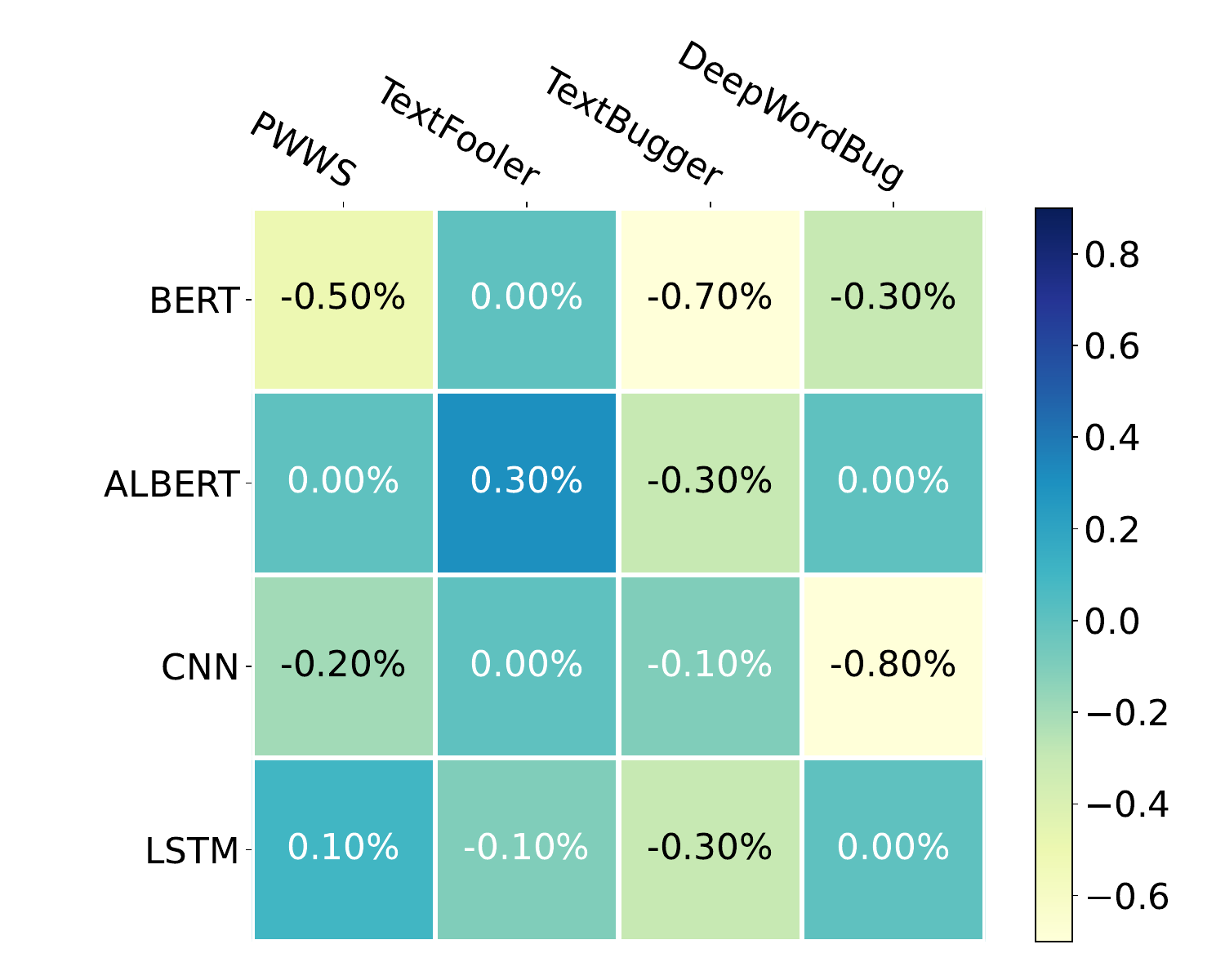}}
        \subfigure[IMDB]{\includegraphics[width=4.35cm]{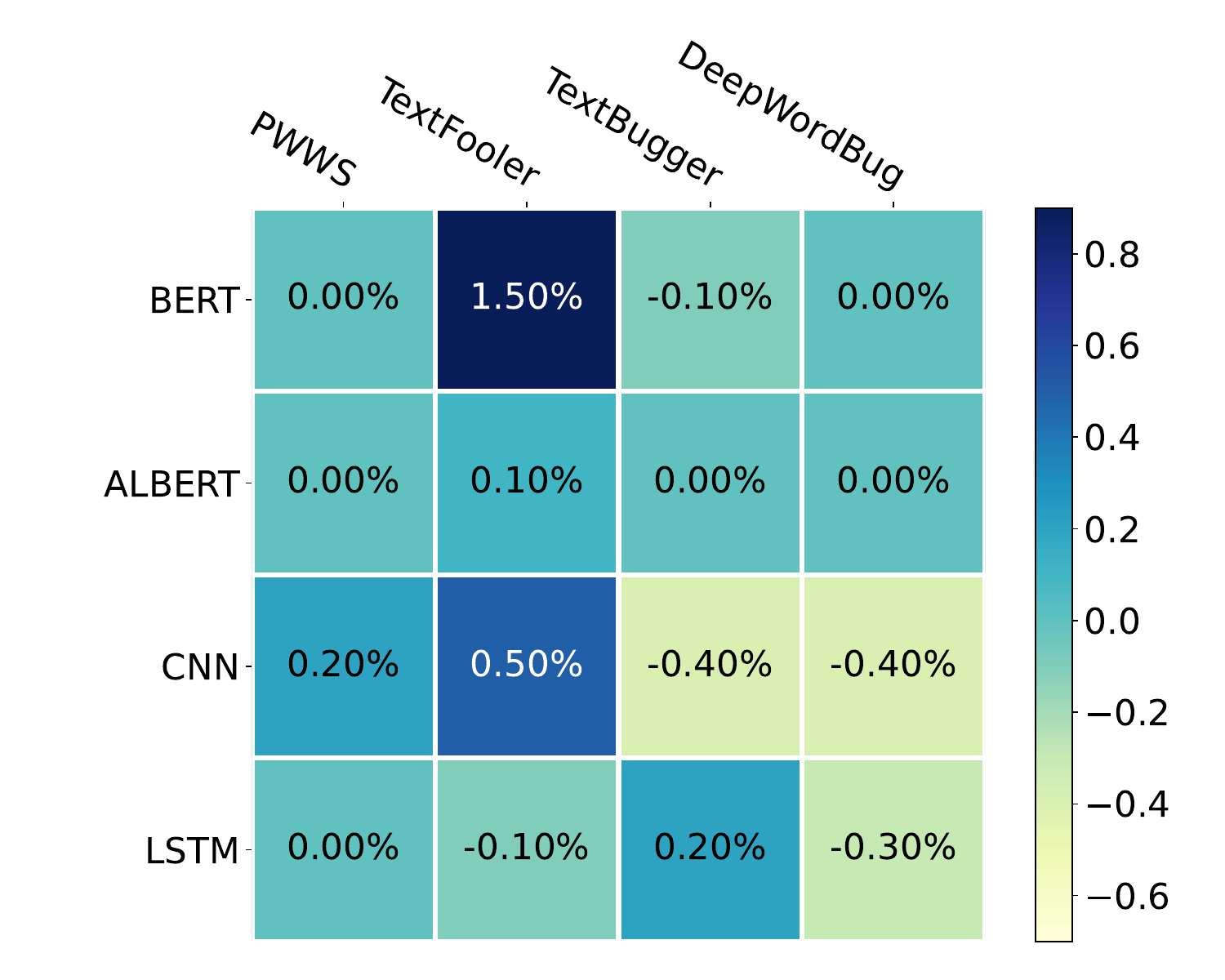}} 
        
       \caption{The F1 score comparison between MLMD-O and MLMD across three datasets, four attack methods, and four victim models. 
       } 
\label{Fig:oracle_nonkey_det}
\end{figure}

Finally, we conduct a comparison of detection performance between MLMD-O (incorporating the oracle masking strategy) and MLMD, as depicted in Fig.~\ref{Fig:oracle_nonkey_det}. After filtering out non-keywords, MLMD-O generally exhibits comparable detection ability to MLMD, with occasional slight reductions, at most only $0.9\%$. This demonstrates that optimizing the mask operation is feasible to reduce costs. There are instances where the detection rate shows a slight improvement. In these cases, after mask and unmask operations, the normal example experiences minimal changes in its manifold, resulting in almost all words in the example having $\sum\limits_{j=1}^{k}\mathbb{I}(z(x), z(u_i^j))<=1$. After adopting the oracle masking strategy, the distinguishable score of normal examples will be further reduced. However, the adversarial example exhibits drastic manifold changes after detection operations, with very few words having $\sum\limits_{j=1}^{k}\mathbb{I}(z(x), z(u_i^j))=1$ in the example. Therefore, the oracle masking strategy has a relatively small impact on the final distinguishable score of the input. Consequently, the distributions of the two types of examples become further separated, improving detection outcomes.

\subsubsection{Employing Gradient Signals to Locate Non-keywords}
\label{subsec:gradient}
We now explore the practical implementation of the oracle method. 
We begin by examining the link between adversarial attacks and gradient signals
\cite{Shen2023TextShieldBS}. The definition of an adversarial example outlined in Sec. \ref{subsec:Notations} can be re-written as:
\begin{equation}
    \underset{\hat{x}}{\arg\max}~ \mathcal L(f(\hat{x}),z(x)),
    \label{Eq:adv_loss}
\end{equation}
where $f(\hat{x})$ is the prediction confidence, and $\mathcal L$ is a loss function. The result of a successful attack is $z(\hat{x}) \neq z(x)$. 
In general, Eq.~(\ref{Eq:adv_loss}) is optimized by a gradient descent method and each adversarial update with a step rate $\epsilon$ for a word can be defined as follows:
\begin{equation}
    e_t^{'}=e_t -\epsilon \dfrac{-\partial \mathcal{L}(f(x),z(x))}{\partial e_t}, 
    \label{Eq:adv_update1}
\end{equation}
where $e_t$ is the word embedding of the word $w_t$ from $x$. 
Many gradient-based attribution works 
\cite{Shen2023TextShieldBS, sundararajan2017axiomatic, Li2015VisualizingAU} suggest that $\Vert \dfrac{-\partial \mathcal{L}(f(x),z(x))}{\partial e_t}\Vert_2$ can be used to measure the importance of input component $e_t$ to the model output. 
Higher gradient values indicate an increased focus on adversarial optimization on specific words. This is because altering these words is more likely to change the predictions of the victim model, thereby achieving the attack objective. Consequently, these words should receive heightened attention in adversarial detection.

Based on this insight, a gradient-guided method is applied to locate non-keywords. We compute the gradient of the loss function $\mathcal{L}$ with respect to $w_t$, producing the importance score $I_t$ to gauge the importance of words during detection:
\begin{equation}
    I_t= \Vert \nabla _ {e_t}{\mathcal L(f(x),z(x))}\Vert_2.
    \label{Eq:important_score}
\end{equation}

We note that the importance score for all words in the input can be efficiently produced by a single backpropagation process. We then arrange words in the input in ascending order according to their importance scores.
The first $ \lfloor(1-r)\times n \rfloor $ words constitute a gradient-guided non-keyword set $G$.
The remaining words are considered keywords, which will be processed while detecting, resulting in a gradient-guided masking strategy.
Substituting MLMD's masking strategy with this gradient-guided approach yields GradMLMD.

GradMLMD leads to substantial reductions in visits to both the masked language model and the victim model by limiting the process to only the last $ \lceil r \times n \rceil$ words from the sorted input. 
To verify the efficacy of GradMLMD, we assess the overlap between non-keywords identified by the oracle method and those identified by the gradient-guided method. 
Similar to 
Sec.~\ref{subsubsec: oracle_method}, 
normal examples and their adversarial counterparts will be separately processed using $\tilde{lap}=\frac{1} {|\tilde{\mathcal{D}}|}  \sum\limits_{i=1}^{|\tilde{\mathcal{D}}|}\dfrac{|\tilde{O}_i \cap \tilde{G}_i |}{| \tilde{G}_i|}$ and $\hat{lap}=\frac{1} {|\hat{\mathcal{D}}|} \sum\limits_{i=1}^{|\hat{\mathcal{D}}|}\dfrac{|\hat{O}_i \cap \hat{G}_i |}{| \hat{G}_i|}$, where $\tilde{O}_i$ and $\tilde{G}_i$ denote the non-keyword sets for a normal text $\tilde{x}_i$ found by the oracle method ($\gamma=1$) and the gradient-guided method, respectively. We then use $\min(\tilde{lap},\hat{lap})$  as the final rate. 
As shown in Fig.~\ref{Fig:grad_orcle_nonkey_bert_SUM1}, the average overlap rate across $12$ settings reaches  $0.86$, which demonstrates the effectiveness of 
using gradient information to locate non-keywords.

\section{Experimental Setup}
\subsection{Datasets and Victim Models}
\label{Subsec:datasets}

We evaluate the detection capabilities of GradMLMD and MLMD using three datasets: AG-NEWS \cite{Zhang2015CharacterlevelCN}, IMDB \cite{Maas2011LearningWV}, and SST \cite{Socher2013RecursiveDM} (see Supplementary Document A for details). The detectors are tested on four widely used victim models: CNN \cite{Zhang2015CharacterlevelCN}, LSTM \cite{Hochreiter1997LongSM}, BERT \cite{Devlin2019BERTPO}, and ALBERT \cite{Lan2020ALBERTAL}, all available in the TextAttack library \cite{Morris2020TextAttackAF}.

\subsection{Attack Methods and Compared Detectors }
\label{Subsec:advAttacks}

We use the open-source toolkit TextAttack to evaluate detectors against four adversarial attacks: PWWS \cite{Ren2019GeneratingNL}, TextFooler \cite{Jin2020IsBR}, TextBugger \cite{Li2019TextBuggerGA}, and DeepWordBug \cite{Gao2018BlackBoxGO}. MLMD/GradMLMD is compared with three state-of-the-art adversarial detectors: FGWS \cite{Mozes2021FrequencyGuidedWS}, WDR \cite{Mosca2022ThatIA}, and GRADMASK \cite{Moon2022GradMaskGT}. Supplementary Document B and C provide further details.

\subsection{Implementation Details}
We collect $1,000$ examples for each combination of dataset ($3$ in total) and victim model architecture ($4$ in total), consisting of 500 normal examples from the test set and 500 adversarial examples generated using four attack algorithms. By default, RoBERTa is used as the masked language model $\Phi$ in MLMD and GradMLMD, with a masking rate $r$ of $1$ for one-by-one masking strategy and $0.3$ for gradient-guided masking. The number of reconstructed texts $k$ is set to $3$. Our experiments show that a basic three-layer MLP suffices for the model-based classifier. We also explore XGBoost as the architecture for the adversarial classifier. While both model- and threshold-based classifiers perform well, we default to the latter unless otherwise specified.


\section{Main Results of MLMD}
\subsection{Detection Performance}
\label{subsec:detect_perf_mlmd}
\begin{figure*}[t]
    \centering 
        \subfigure[PWWS]{\includegraphics[width=4.25cm]{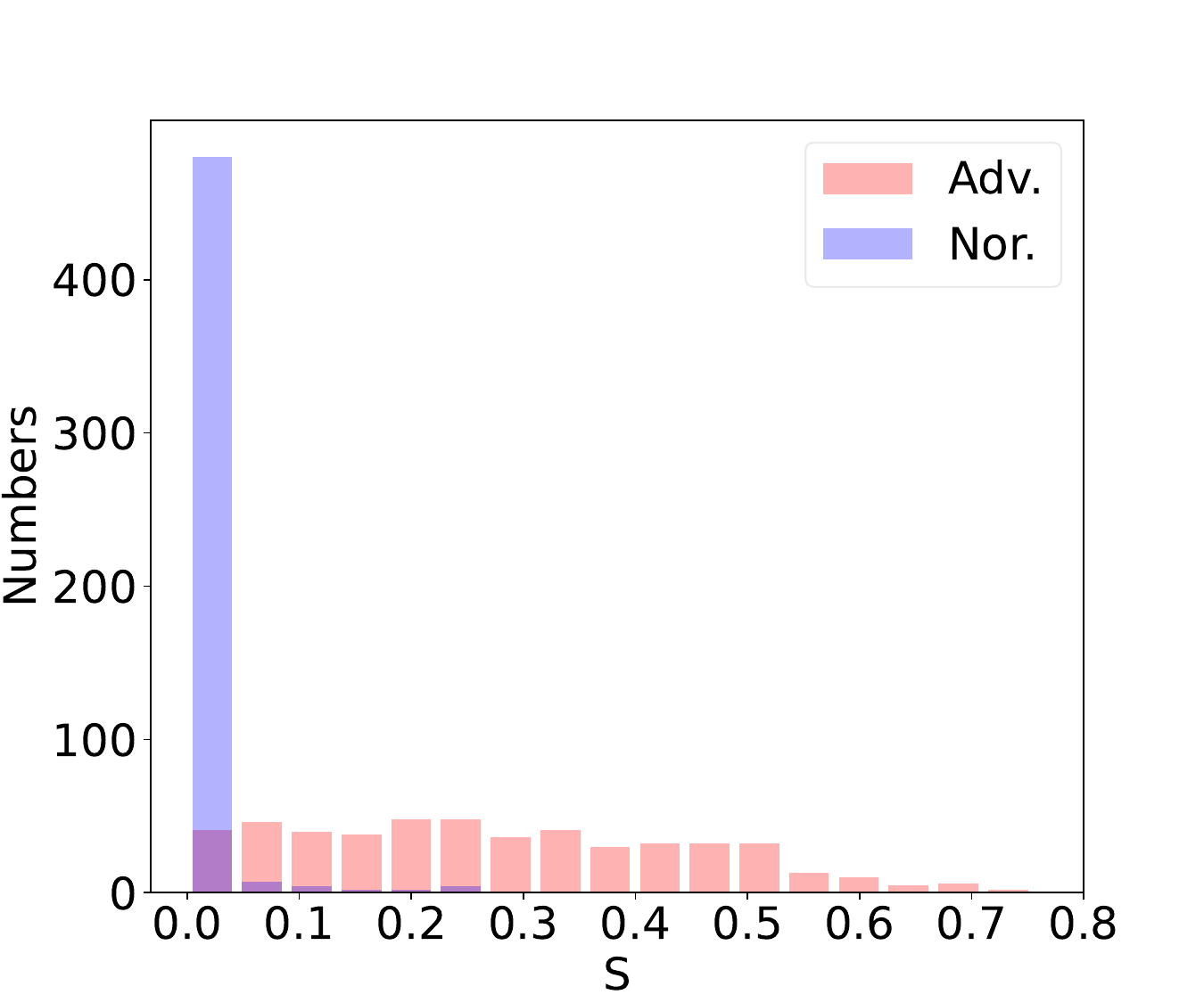}}
        \subfigure[TextFooler]{\includegraphics[width=4.25cm]{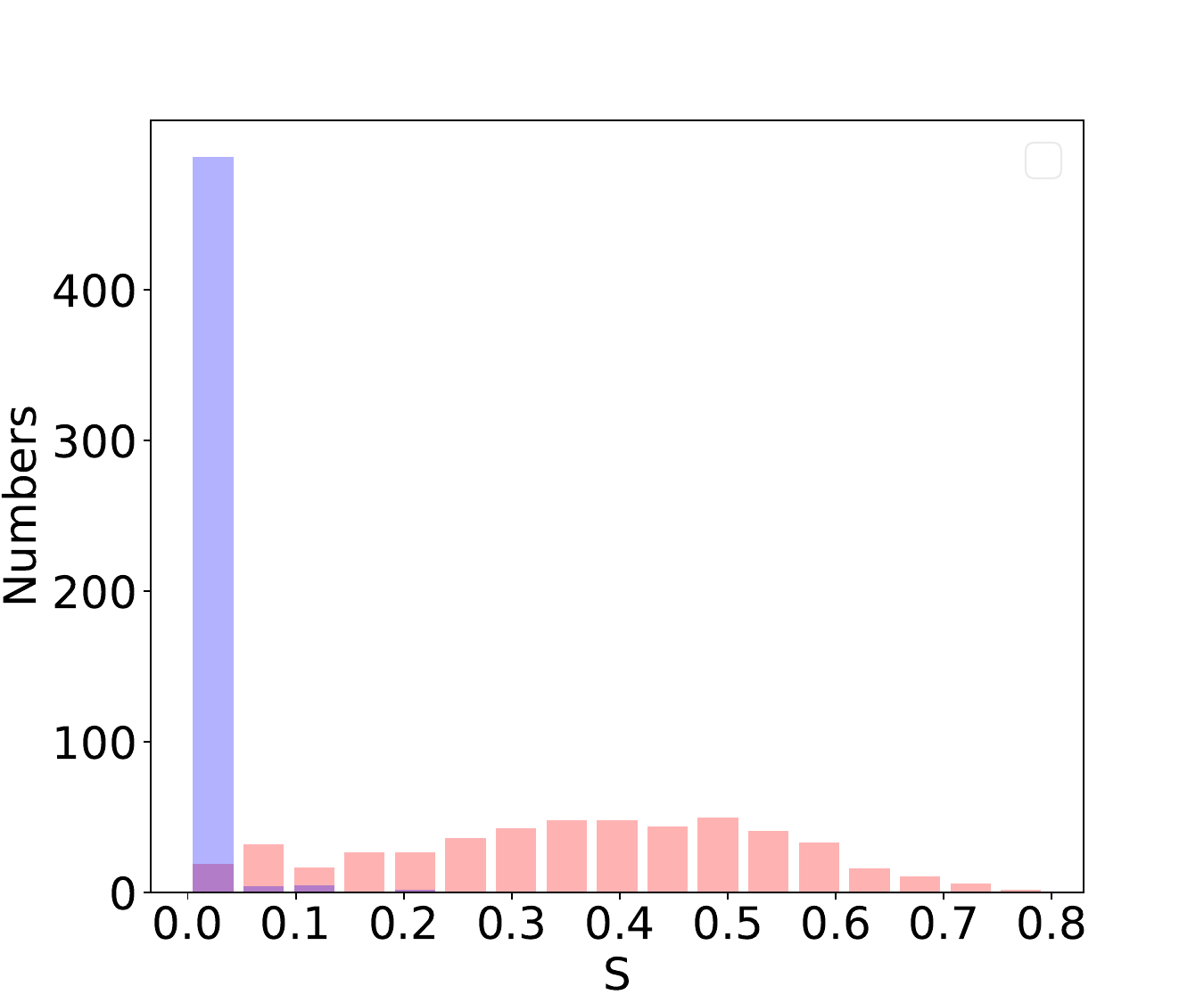}} 
        \subfigure[TextBugger]{\includegraphics[width=4.25cm]{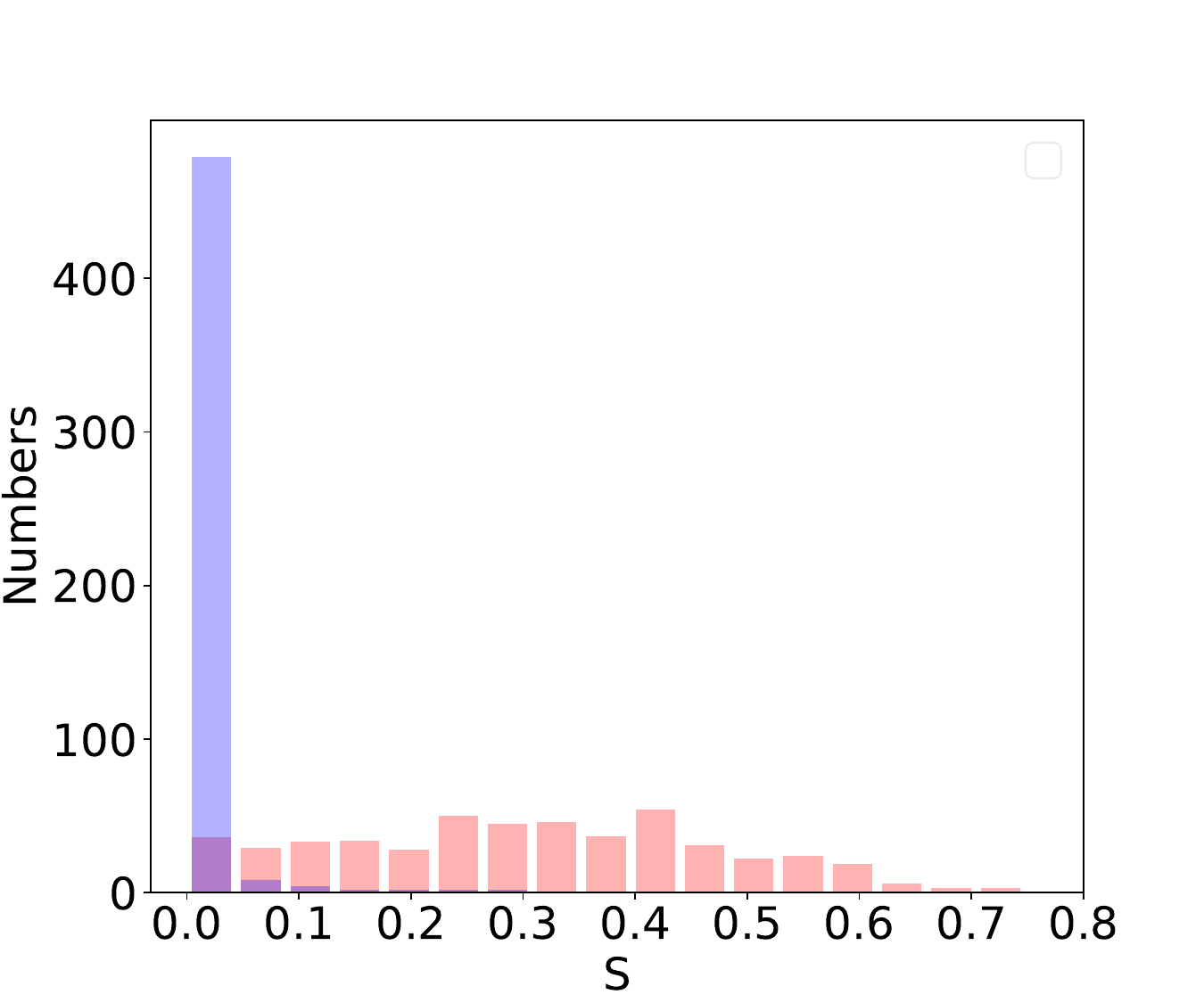}}
        \subfigure[DeepWordBug]{\includegraphics[width=4.25cm]{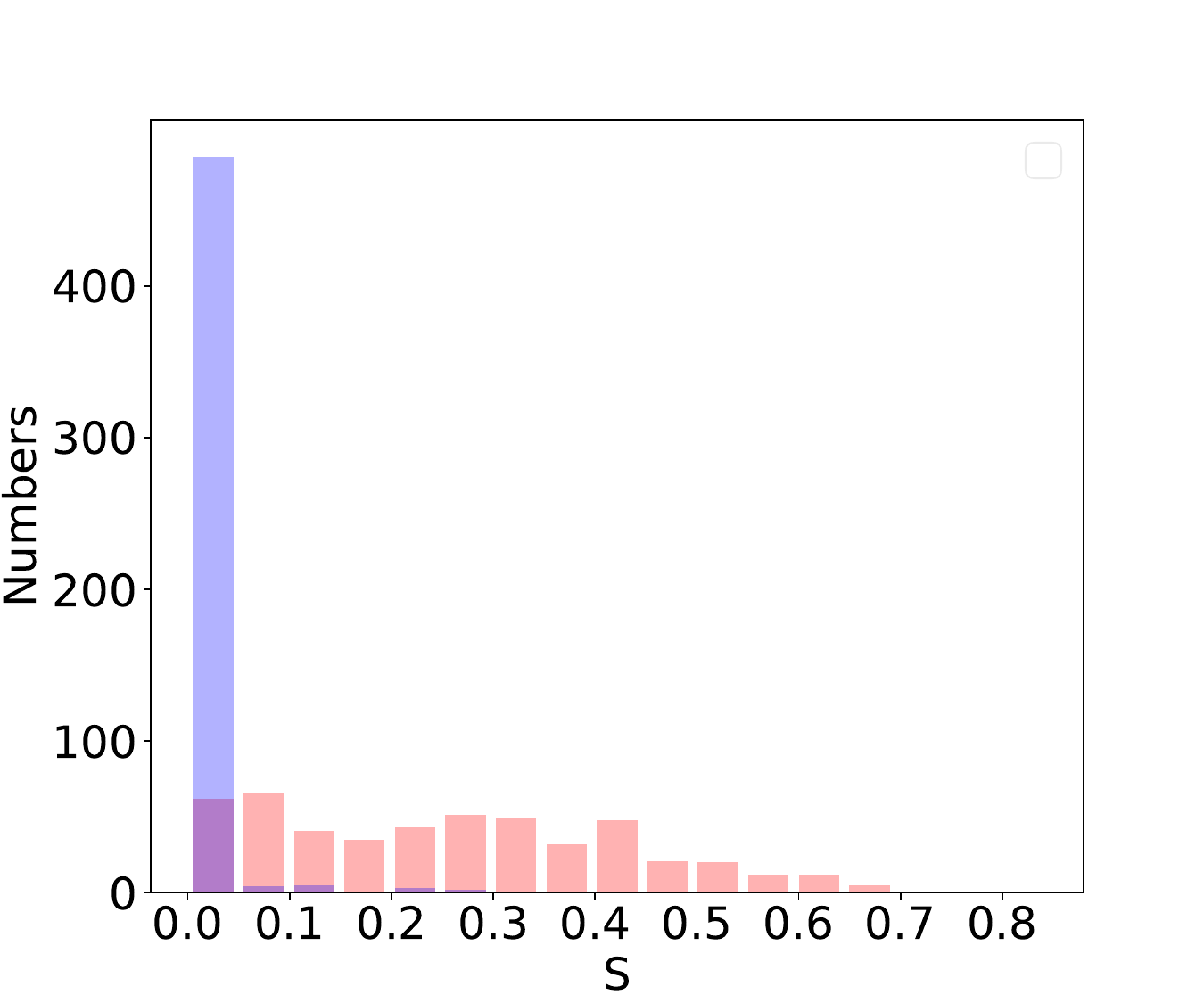}}
        \vspace{-0.2cm}
        \caption{The histogram displays the distinguishable scores $S$ defined by Eq.~(\ref{Eq:disScores}) calculated for normal examples and their corresponding adversarial counterparts generated by attacking BERT trained on AG-NEWS using four different attack methods. Adversarial examples exhibit substantial changes in manifold following the mask and unmask operations, resulting in significantly different predictions (of the victim model) from  original inputs. In contrast, after the two detection operations, normal inputs are  still projected back to the manifold of normal data, ensuring consistency in their prediction results with the original ones. Thus, their distinguishable scores $S$ tend to cluster tightly around $0.0$.}
\label{Fig:D_scores}
\end{figure*}

The primary detection results can be found in Table \ref{tab:detect_perf_mlmd} and Table \ref{tab:de_pre_sst2} (Supplementary Document D). Significantly, our initial method, MLMD, demonstrates substantial superiority over competing approaches across various scenarios, achieving an average F1 score improvement ranging from $1\%$ to $20\%$. These results highlight MLMD's model-agnostic and attack-agnostic capabilities.
A significant limitation of prior approaches employing special token or synonym substitution techniques is their tendency to cause a significant decrease in task accuracy on non-adversarial examples. 
However, our method excels in adversarial detection without compromising overall task accuracy. This result provides empirical validation of the benefits associated with employing a masked language model for adversarial detection. The performance improvement indicates that the changes in manifold resulting from the introduction of a masked language model and MLM objective are more effective in distinguishing adversarial inputs than the changes induced by replacing synonyms or special tokens.

This observation can be further verified by visualizing the distinguishable score defined by Eq.~(\ref{Eq:disScores}). In Fig.~\ref{Fig:D_scores}, a clear divergence emerges between normal and adversarial examples. 
Notably, the distinguishable scores for normal examples converge around $0$, whereas scores for adversarial texts are scattered across values exceeding $0$. This emphasizes the masked language models' role in maintaining the benign nature of normal examples while inducing manifold changes in adversarial examples, resulting in clearly distinguishable signals.

GRADMASK appears to depend heavily on the capabilities of the victim model. 
For example, as transformer-based victim models often exhibit a high confidence score for the predicted class, 
when input is adversarial, GRADMASK is able to introduce a difference in the confidence score of the same class before and after the mask operation. However, CNN and LSTM models typically assign more moderate scores to the predicted label. Consequently, after applying GRADMASK's mask operation, it becomes challenging to identify an appropriate threshold to distinguish between adversarial and normal texts due to their subtle differences.


Additionally, TextFooler is relatively easy to detect due to its strategy of crafting adversarial examples through precise word substitutions with similar embeddings. This method keeps perturbed examples close to the decision boundary, resulting in noticeable changes in the manifold of these examples when masked and unmasked, making them easier to detect.

\begin{table}[htbp]
  \centering
  \caption{Detection performance of FGWS, WDR, GRADMASK, and MLMD on AG-NEWS and IMDB. We omit the detection results of FGWS for adversarial examples generated by TextBugger attack and DeepWordBug attack, as it struggles to locate appropriate synonyms from training sets for certain words when only characters are perturbed.}
  \resizebox{0.48\textwidth}{!}{
    \begin{tabular}{ccl|cc|cc|cc|cc}
    \toprule
    \multirow{2}[4]{*}{\textbf{Dataset}} & \multirow{2}[4]{*}{\textbf{Model}} & \multicolumn{1}{l}{\multirow{2}[4]{*}{\textbf{Method}}} & \multicolumn{2}{c}{\textbf{PWWS}} & \multicolumn{2}{c}{\textbf{TextFooler}} & \multicolumn{2}{c}{\textbf{TextBugger}} & \multicolumn{2}{c}{\textbf{DeepWordBug}} \\
\cmidrule{4-11}          &       & \multicolumn{1}{c}{} & \textbf{Acc.} & \multicolumn{1}{c}{\textbf{F1}} & \textbf{Acc.} & \multicolumn{1}{c}{\textbf{F1}} & \textbf{Acc.} & \multicolumn{1}{c}{\textbf{F1}} & \textbf{Acc.} & \textbf{F1} \\
    \midrule
    \multicolumn{1}{c}{\multirow{16}[8]{*}{\makecell{AG-\\NE\\WS}}} & \multirow{4}[2]{*}{BERT} & FGWS  & 0.891  & 0.885  & 0.878  & 0.868  &   -    &   -    &   -    &   -    \\
          &       & WDR   & \cellcolor[rgb]{ .988,  .894,  .839}0.964  & 0.959  & 0.970  & 0.971  & 0.937  & 0.934  & 0.907  & 0.903  \\
          &       & GRADMASK & 0.953  & 0.955  & 0.962  & 0.969  & 0.890  & 0.907  & 0.894  & 0.903  \\
          &       & MLMD  & 0.959  & \cellcolor[rgb]{ .988,  .894,  .839}0.961  & \cellcolor[rgb]{ .988,  .894,  .839}0.983  & \cellcolor[rgb]{ .988,  .894,  .839}0.985  & \cellcolor[rgb]{ .988,  .894,  .839}0.950  & \cellcolor[rgb]{ .988,  .894,  .839}0.950  & \cellcolor[rgb]{ .988,  .894,  .839}0.938  & \cellcolor[rgb]{ .988,  .894,  .839}0.940  \\
\cmidrule{2-11}          & \multirow{4}[2]{*}{ALBERT} & FGWS  & 0.885  & 0.878  & 0.864  & 0.850  &   -    &   -    &   -    &   -    \\
          &       & WDR   & 0.931  & 0.929  & 0.951  & 0.940  & 0.938  & 0.931  & 0.891  & 0.897  \\
          &       & GRADMASK & 0.905  & 0.908  & 0.938  & 0.940  & 0.905  & 0.906  & 0.892  & 0.894  \\
          &       & MLMD  & \cellcolor[rgb]{ .988,  .894,  .839}0.952  & \cellcolor[rgb]{ .988,  .894,  .839}0.950  & \cellcolor[rgb]{ .988,  .894,  .839}0.984  & \cellcolor[rgb]{ .988,  .894,  .839}0.984  & \cellcolor[rgb]{ .988,  .894,  .839}0.965  & \cellcolor[rgb]{ .988,  .894,  .839}0.965  & \cellcolor[rgb]{ .988,  .894,  .839}0.943  & \cellcolor[rgb]{ .988,  .894,  .839}0.943  \\
\cmidrule{2-11}          & \multirow{4}[2]{*}{CNN} & FGWS  & 0.900  & 0.895  & 0.813  & 0.783  &   -    &   -    &   -    &   -    \\
          &       & WDR   & 0.913  & 0.911  & 0.934  & 0.937  & 0.909  & 0.916  & 0.903  & 0.912  \\
          &       & GRADMASK & 0.806  & 0.836  & 0.800  & 0.803  & 0.758  & 0.802  & 0.770  & 0.757  \\
          &       & MLMD  & \cellcolor[rgb]{ .988,  .894,  .839}0.964  & \cellcolor[rgb]{ .988,  .894,  .839}0.965  & \cellcolor[rgb]{ .988,  .894,  .839}0.971  & \cellcolor[rgb]{ .988,  .894,  .839}0.972  & \cellcolor[rgb]{ .988,  .894,  .839}0.957  & \cellcolor[rgb]{ .988,  .894,  .839}0.958  & \cellcolor[rgb]{ .988,  .894,  .839}0.956  & \cellcolor[rgb]{ .988,  .894,  .839}0.957  \\
\cmidrule{2-11}          & \multirow{4}[2]{*}{LSTM} & FGWS  & 0.871  & 0.862  & 0.807  & 0.776  &   -    &   -    &   -    &   -    \\
          &       & WDR   & 0.910  & 0.907  & 0.932  & 0.930  & 0.893  & 0.885  & 0.904  & 0.910  \\
          &       & GRADMASK & 0.813  & 0.860  & 0.828  & 0.863  & 0.810  & 0.824  & 0.800  & 0.798  \\
          &       & MLMD  & \cellcolor[rgb]{ .988,  .894,  .839}0.954  & \cellcolor[rgb]{ .988,  .894,  .839}0.955  & \cellcolor[rgb]{ .988,  .894,  .839}0.969  & \cellcolor[rgb]{ .988,  .894,  .839}0.967  & \cellcolor[rgb]{ .988,  .894,  .839}0.949  & \cellcolor[rgb]{ .988,  .894,  .839}0.949  & \cellcolor[rgb]{ .988,  .894,  .839}0.942  & \cellcolor[rgb]{ .988,  .894,  .839}0.946  \\
    \midrule
    \multicolumn{1}{c}{\multirow{16}[8]{*}{\makecell{IM\\DB}}} & \multirow{4}[2]{*}{BERT} & FGWS  & 0.892  & 0.881  & 0.880  & 0.867  &   -    &   -    &   -    &   -    \\
          &       & WDR   & \cellcolor[rgb]{ .988,  .894,  .839}0.946  & \cellcolor[rgb]{ .988,  .894,  .839}0.950  & \cellcolor[rgb]{ .988,  .894,  .839}0.948  & \cellcolor[rgb]{ .988,  .894,  .839}0.949  & 0.950  & 0.946  & 0.915  & 0.909  \\
          &       & GRADMASK & 0.943  & 0.943  & 0.922  & 0.925  & 0.907  & 0.911  & 0.870  & 0.875  \\
          &       & MLMD  & 0.943  & 0.945  & 0.946  & 0.947  & \cellcolor[rgb]{ .988,  .894,  .839}0.955  & \cellcolor[rgb]{ .988,  .894,  .839}0.956  & \cellcolor[rgb]{ .988,  .894,  .839}0.934  & \cellcolor[rgb]{ .988,  .894,  .839}0.936  \\
\cmidrule{2-11}          & \multirow{4}[2]{*}{ALBERT} & FGWS  & 0.712  & 0.649  & 0.822  & 0.814  &   -    &   -    &   -    &   -    \\
          &       & WDR   & 0.801  & 0.807  & 0.873  & 0.882  & 0.882  & 0.893  & 0.799  & 0.811  \\
          &       & GRADMASK & 0.945  & \cellcolor[rgb]{ .988,  .894,  .839}0.951  & \cellcolor[rgb]{ .988,  .894,  .839}0.968  & \cellcolor[rgb]{ .988,  .894,  .839}0.973  & 0.895  & 0.901  & 0.876  & 0.883  \\
          &       & MLMD  & \cellcolor[rgb]{ .988,  .894,  .839}0.947  & 0.950  & 0.967  & 0.970  & \cellcolor[rgb]{ .988,  .894,  .839}0.944  & \cellcolor[rgb]{ .988,  .894,  .839}0.951  & \cellcolor[rgb]{ .988,  .894,  .839}0.893  & \cellcolor[rgb]{ .988,  .894,  .839}0.895  \\
\cmidrule{2-11}          & \multirow{4}[2]{*}{CNN} & FGWS  & \cellcolor[rgb]{ .988,  .894,  .839}0.903  & \cellcolor[rgb]{ .988,  .894,  .839}0.905  & 0.764  & 0.705  &   -    &   -    &   -    &   -    \\
          &       & WDR   & 0.845  & 0.851  & 0.872  & 0.880  & 0.844  & 0.860  & 0.820  & 0.835  \\
          &       & GRADMASK & 0.794  & 0.789  & 0.750  & 0.755  & 0.721  & 0.766  & 0.706  & 0.732  \\
          &       & MLMD  & 0.898  & 0.903  & \cellcolor[rgb]{ .988,  .894,  .839}0.927  & \cellcolor[rgb]{ .988,  .894,  .839}0.929  & \cellcolor[rgb]{ .988,  .894,  .839}0.915  & \cellcolor[rgb]{ .988,  .894,  .839}0.922  & \cellcolor[rgb]{ .988,  .894,  .839}0.903  & \cellcolor[rgb]{ .988,  .894,  .839}0.908  \\
\cmidrule{2-11}          & \multirow{4}[2]{*}{LSTM} & FGWS  & 0.801  & 0.823  & 0.718  & 0.644  &   -    &   -    &   -    &   -    \\
          &       & WDR   & 0.841  & 0.843  & 0.864  & 0.869  & 0.857  & 0.862  & 0.831  & 0.833  \\
          &       & GRADMASK & 0.801  & 0.789  & 0.803  & 0.788  & 0.779  & 0.773  & 0.775  & 0.834  \\
          &       & MLMD  & \cellcolor[rgb]{ .988,  .894,  .839}0.886  & \cellcolor[rgb]{ .988,  .894,  .839}0.890  & \cellcolor[rgb]{ .988,  .894,  .839}0.906  & \cellcolor[rgb]{ .988,  .894,  .839}0.918  & \cellcolor[rgb]{ .988,  .894,  .839}0.900  & \cellcolor[rgb]{ .988,  .894,  .839}0.902  & \cellcolor[rgb]{ .988,  .894,  .839}0.895  & \cellcolor[rgb]{ .988,  .894,  .839}0.906  \\
    \bottomrule
    \end{tabular}%
    }
  \label{tab:detect_perf_mlmd}%
\end{table}%

\subsection{The Detection Performance of the Model-based Classifier}
\label{Sec:model_based_performance}

Table \ref{tab:model_pre} in Supplementary Document E illustrates MLMD's performance with model-based classifiers. 
In general, the performance of model-based classifiers closely aligns with that of threshold-based classifiers, regardless of the model architecture. This suggests that both normal and adversarial examples exhibit
distinguishable signals through mask and unmask operations, easily separable via either thresholding
or classifiers with varied model architectures.
In addition, the performance improves when trained on the dataset $\overline{\Gamma}$ compared to $\Gamma$.
This indicates that adding information about the order of elements will greatly assist the model-based method in extracting features to recognize adversarial inputs.



\section{Main Results of GradMLMD}
\subsection{Detection Performance}
\label{subsec:detect_perf_gradmlmd}

\begin{figure}[htbp]
  \centering
  \resizebox{0.5\textwidth}{!}{
    \begin{tabular}{ccc}
    \centering
     \subfigure{ \includegraphics[width=4.9cm]{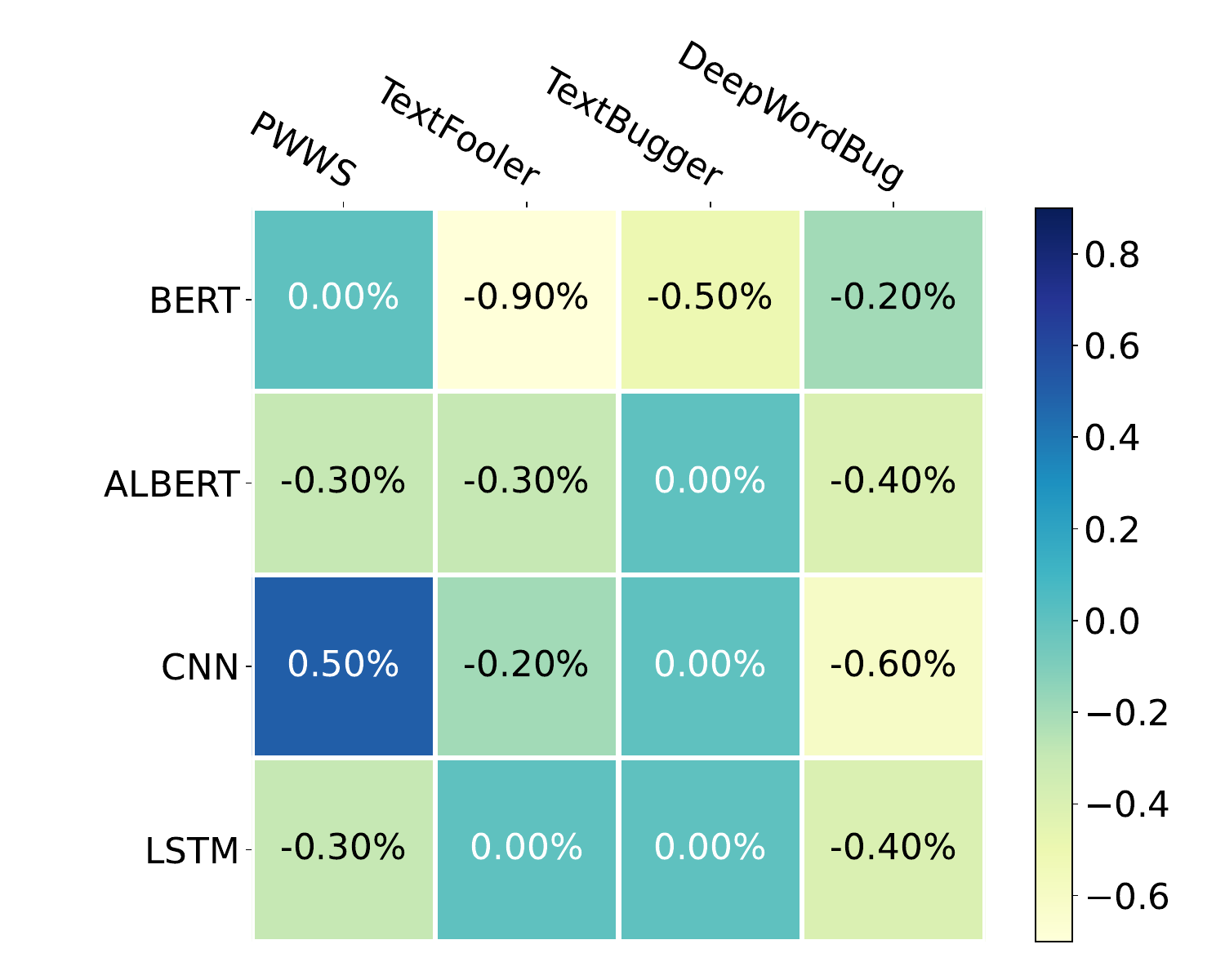}} & \subfigure{\includegraphics[width=4.6cm]{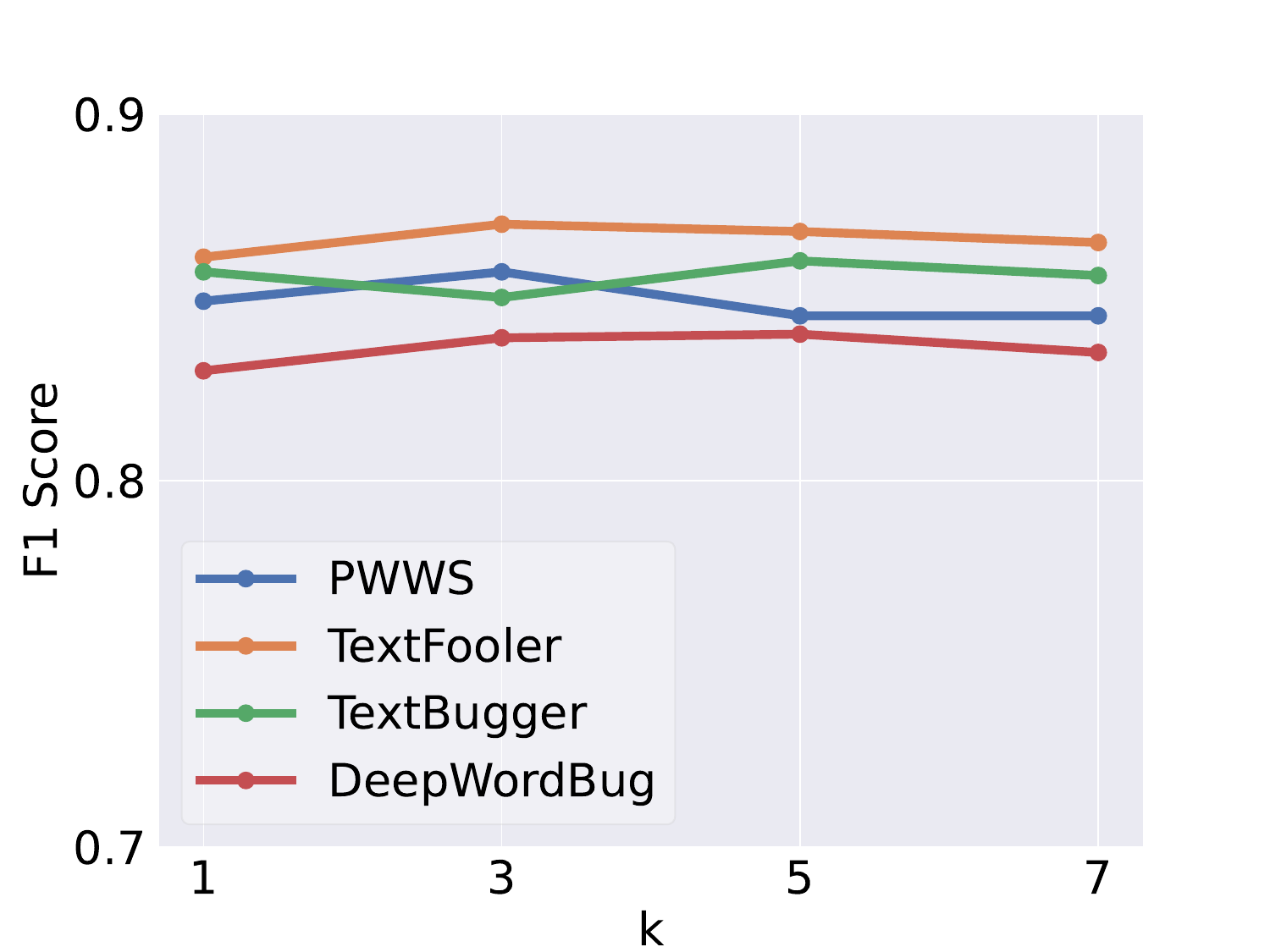}} \\
    \small{(a) SST-2} & \small{(d) SST-2} \\
     \subfigure{\includegraphics[width=4.9cm]{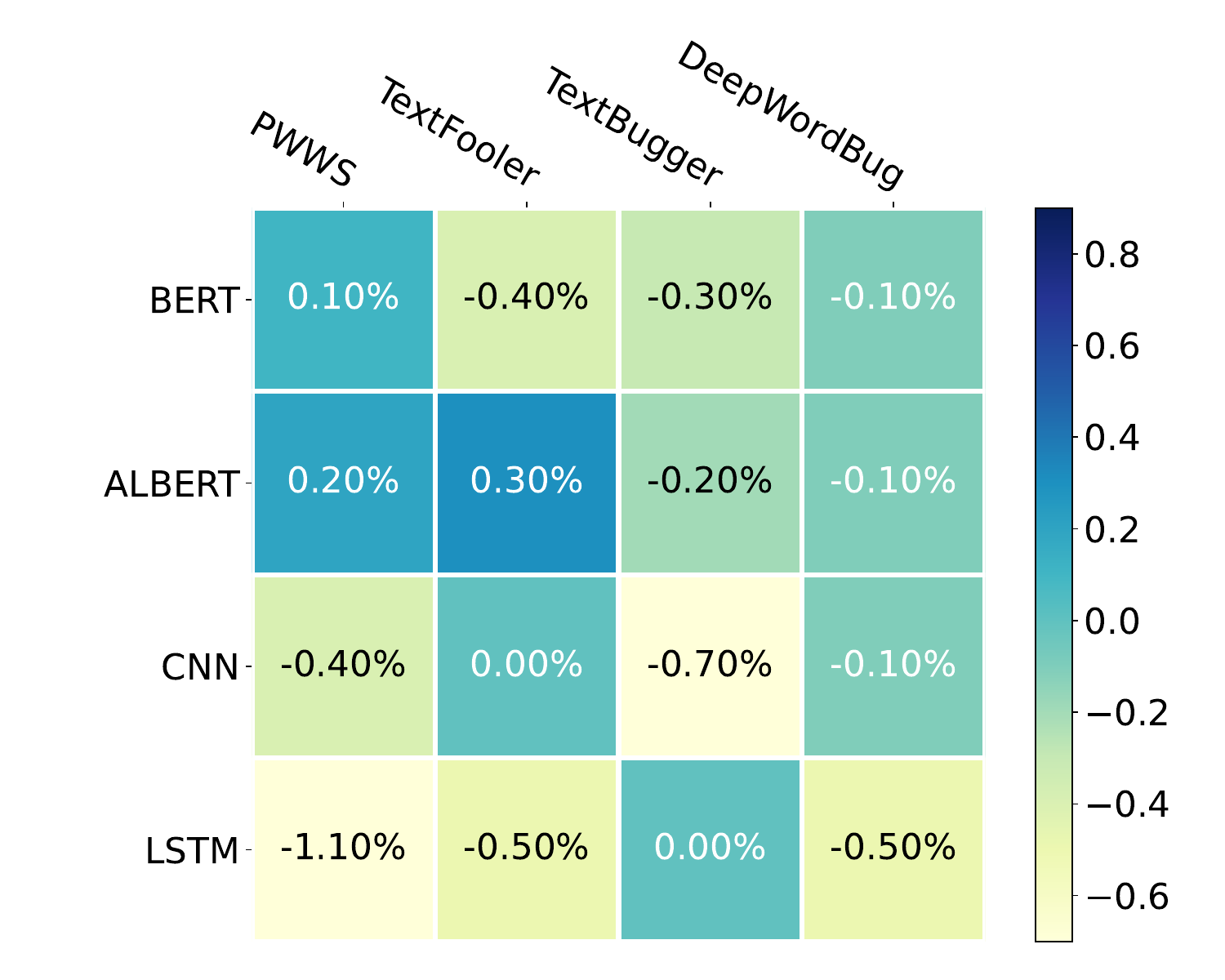}} & \subfigure{\includegraphics[width=4.6cm]{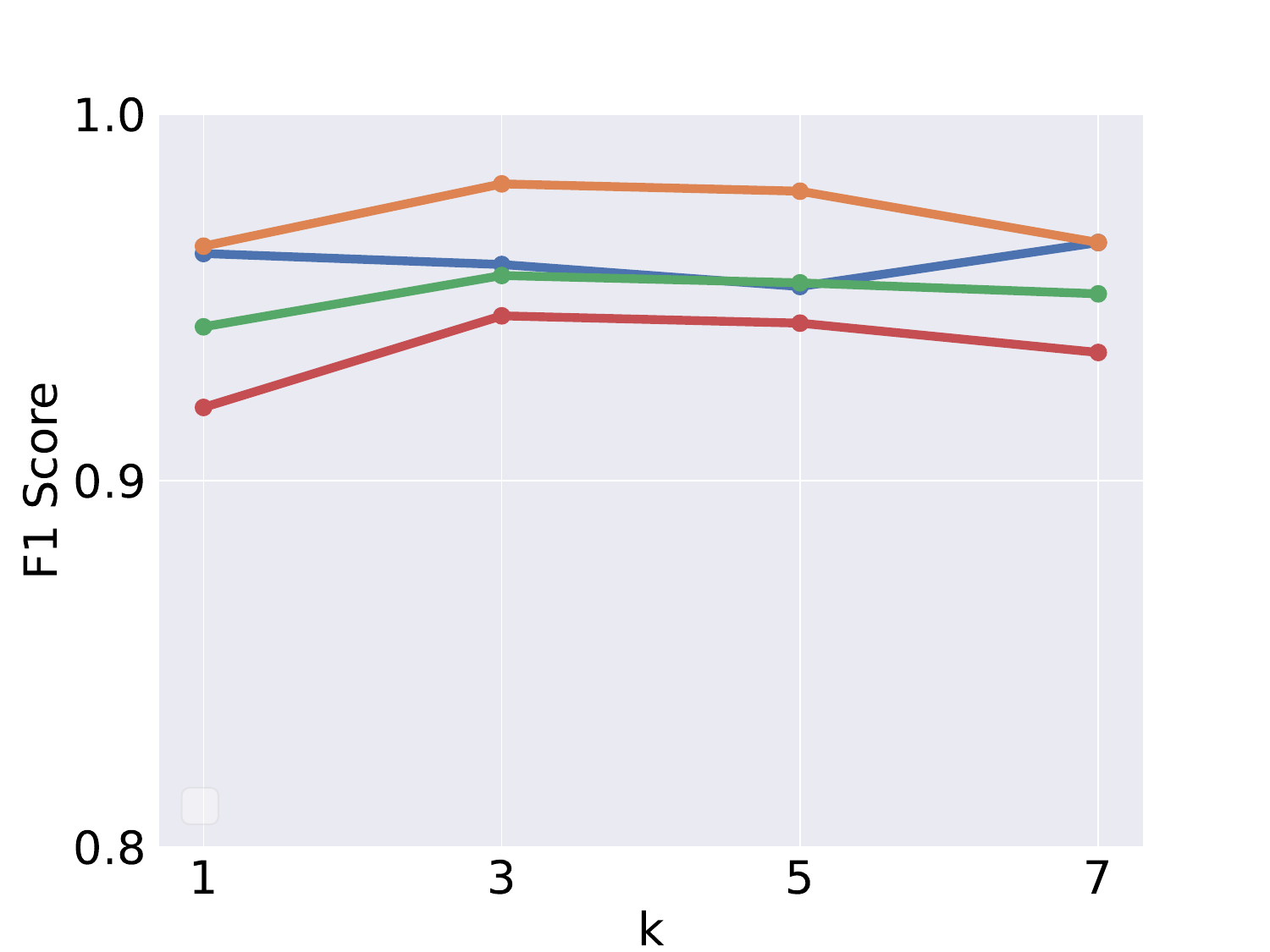}} \\
    \small{(b) AG-NEWS} & \small{(e) AG-NEWS} \\
    \subfigure{\includegraphics[width=4.9cm]{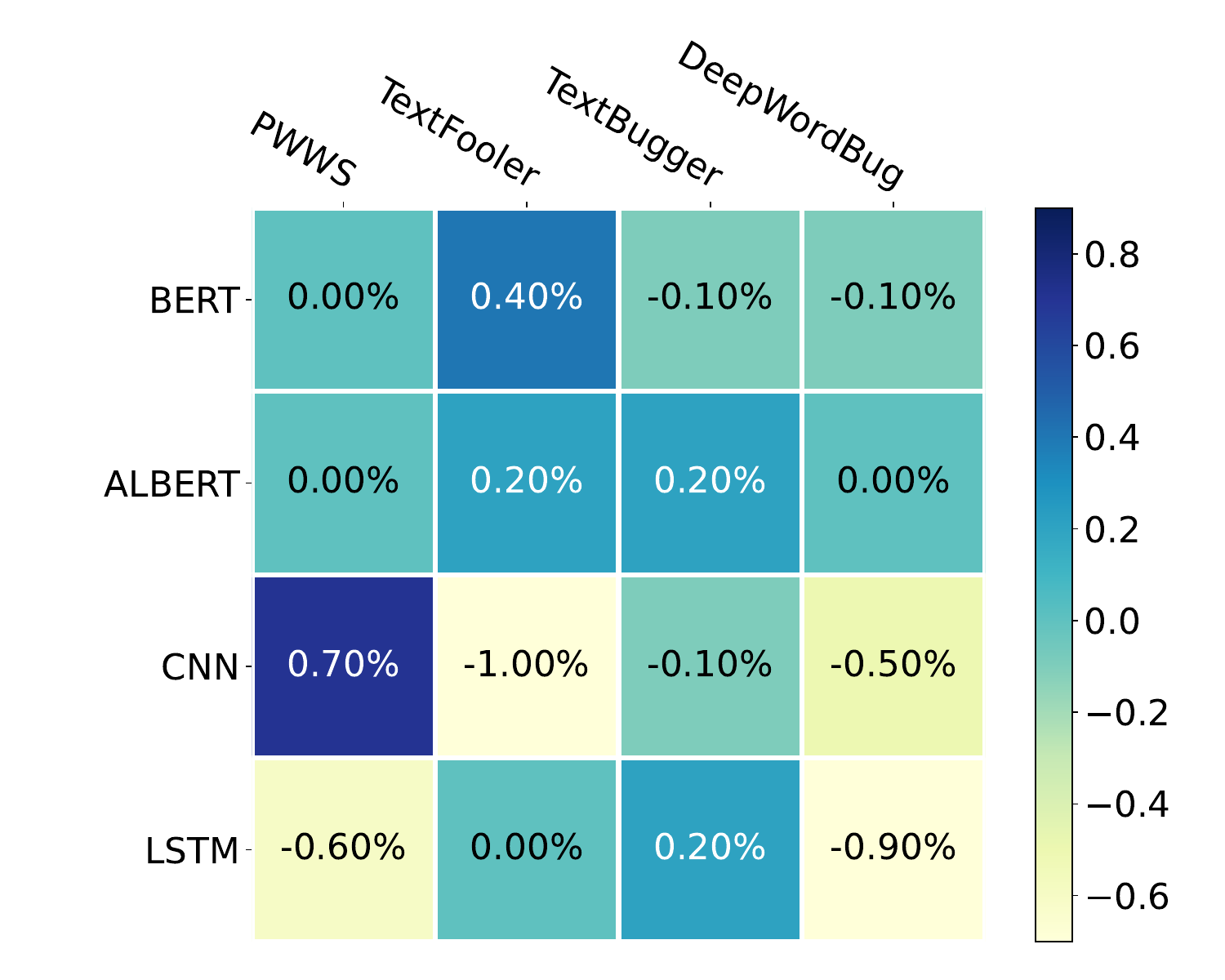}} & \subfigure{\includegraphics[width=4.6cm]{pic/topk\_imdb\_bert\_modify.pdf}}  \\
    \small{(c) IMDB} & \small{(f) IMDB} \\
    \end{tabular}%
    }
    \caption{(a-c) Comparisons of the detection performance in terms of F1 score between MLMD and GradMLMD across three datasets, four attack methods, and four victim models. The results show that GradMLMD has a comparable ability to detect adversarial examples as the original MLMD. (d-f) The impact of the number of reconstruction times $k$ in the unmask operation on the detection performance of GradMLMD. The victim model is the fine-tuned BERT.}
  \label{Fig:gradmlmd_differences}%
\end{figure}%

        

We compute the differences in detection performance in terms of F1 score between MLMD and GradMLMD. 
The results (Fig.~\ref{Fig:gradmlmd_differences}(a-c)) show that across all experimental configurations, most of GradMLMD's detection results demonstrate comparable performance to the original MLMD, with declines, if any, not exceeding $0.5\%$. However, there are instances where detection shows a noticeable decrease in a few cases ($5$ out of $48$ settings). 
For instance, when detecting adversarial examples generated by using PWWS attack the LSTM, which is fine-tuned on AG-NEWS, GradMLMD's score is $1.1\%$ lower than MLMD. It is noteworthy that even in such cases, GradMLMD still outperforms the state-of-the-art algorithms (i.e., FGWS, WDR and GRADMASK). 
In some cases, GradMLMD's detection capability even slightly improves upon MLMD. 
Due to the gradient masking strategy being an effective approximation of the oracle strategy, in these cases, this strategy causes the distinguishable scores of normal examples to shift towards $0$ more dramatically than the scores of adversarial examples shift towards smaller values. As a result, 
the further reduction in overlap between the distinguishable score distributions of the two example types enhances detection performance.

In summary, GradMLMD sustains a detection capability comparable to MLMD. This implies that the exclusion of non-keywords has a negligible effect on detection outcomes. The  masking strategy in GradMLMD reduces the mask and unmask operations to $30\%$ of words based on their importance scores, significantly saving the cost caused by the  interactions between the masked language models and victim models. 

\subsection{Exploring Various Masked Language Models for Detection}
\label{subsec:various_models}

In this section, we study how various masked language models influence detection performance. We integrate three widely-used masked language models (BERT, ALBERT, and RoBERTa) into GradMLMD.
This evaluation is performed on pairs of normal and adversarial texts crafted by attacking four victim models with the TextFooler attack.

\begin{table}[htbp]
  \centering
  \caption{The impact of various masked language models on the detection results of GradMLMD. We consider three widely-used masked language models: BERT, ALBERT, and RoBERTa.}
  \resizebox{0.40\textwidth}{!}{
    \begin{tabular}{cl|cc|cc|cc}
    \toprule
    \multirow{2}[4]{*}{\textbf{Dataset}} & \multicolumn{1}{c}{\multirow{2}[4]{*}{\textbf{Model}}} & \multicolumn{2}{c}{\textbf{BERT}} & \multicolumn{2}{c}{\textbf{ALBERT}} & \multicolumn{2}{c}{\textbf{RoBERTa}} \\
\cmidrule{3-8}          & \multicolumn{1}{c}{} & \textbf{Acc.}  & \multicolumn{1}{c}{\textbf{F1}} & \textbf{Acc.}  & \multicolumn{1}{c}{\textbf{F1}} &\textbf{Acc.}  & \textbf{F1} \\
    \midrule
    \multirow{4}[2]{*}{SST-2} & BERT  & 0.840 & 0.845 & 0.840 & 0.849 & \cellcolor[rgb]{ .988,  .894,  .839}0.868  & \cellcolor[rgb]{ .988,  .894,  .839}0.870  \\
          & ALBERT & 0.829 & 0.843 & 0.843 & 0.844 & \cellcolor[rgb]{ .988,  .894,  .839}0.853  & \cellcolor[rgb]{ .988,  .894,  .839}0.854  \\
          & CNN   & 0.830 & 0.845 & 0.835 & 0.845 & \cellcolor[rgb]{ .988,  .894,  .839}0.852  & \cellcolor[rgb]{ .988,  .894,  .839}0.857  \\
          & LSTM  & 0.819 & 0.831 & 0.820 & 0.831 & \cellcolor[rgb]{ .988,  .894,  .839}0.832  & \cellcolor[rgb]{ .988,  .894,  .839}0.843  \\
    \midrule
    \multirow{4}[2]{*}{AG-NEWS} & BERT  & 0.961 & 0.962 & 0.965 & 0.966 & \cellcolor[rgb]{ .988,  .894,  .839}0.979  & \cellcolor[rgb]{ .988,  .894,  .839}0.981  \\
          & ALBERT & 0.969 & 0.971 & 0.970 & 0.972 & \cellcolor[rgb]{ .988,  .894,  .839}0.985  & \cellcolor[rgb]{ .988,  .894,  .839}0.987  \\
          & CNN   & 0.963 & 0.966 & 0.965 & 0.966 & \cellcolor[rgb]{ .988,  .894,  .839}0.970  & \cellcolor[rgb]{ .988,  .894,  .839}0.972  \\
          & LSTM  & 0.954 & 0.955 & \cellcolor[rgb]{ .988,  .894,  .839}0.970 & \cellcolor[rgb]{ .988,  .894,  .839}0.969 & 0.960  & 0.962  \\
    \midrule
    \multirow{4}[2]{*}{IMDB} & BERT  & 0.948 & 0.950  & 0.941 & 0.943 & \cellcolor[rgb]{ .988,  .894,  .839}0.950  & \cellcolor[rgb]{ .988,  .894,  .839}0.951  \\
          & ALBERT & 0.925 & 0.924 & 0.949 & 0.950  & \cellcolor[rgb]{ .988,  .894,  .839}0.970  & \cellcolor[rgb]{ .988,  .894,  .839}0.972  \\
          & CNN   & 0.911 & 0.913 & \cellcolor[rgb]{ .988,  .894,  .839}0.922 & \cellcolor[rgb]{ .988,  .894,  .839}0.920 & 0.921  & 0.919  \\
          & LSTM  & 0.901 & 0.905 & 0.900 & 0.903 & \cellcolor[rgb]{ .988,  .894,  .839}0.906  & \cellcolor[rgb]{ .988,  .894,  .839}0.918  \\
    \bottomrule
    \end{tabular}%
    }
  \label{tab:vaious_effect}%
  
\end{table}%

The results in Table~\ref{tab:vaious_effect} show that regardless of which masked language model is used as the component, the corresponding detectors perform well.
This reveals that
a wide range of masked language models can be used for adversarial attack detection within our framework.


In addition, when changing from BERT to RoBERTa as the masked language model component, a stable improvement in detection performance is observed. This stems from the distinct capabilities they possess in approximating the manifold of normal examples. The rationale behind this trend can be attributed to several factors:

\textbf{Pre-training data.} Although BERT and ALBERT are pre-trained on English Wikipedia and Bookcorpus, 
the training data of RoBERTa is significantly larger and  more heterogeneous than that of them. This empowers RoBERTa to learn richer language representations for normal data, thereby enhancing its ability to approximate the manifold of normal examples compared to both BERT and ALBERT.

\textbf{Pre-training task.} These three models are designed differently.
During the mask operation, BERT employs a static masking strategy, meaning each data instance has only one masked copy in training stage. 
Unlike BERT, which handles individual tokens, ALBERT employs an n-gram masking strategy to generate masked text, where each masked position can be composed of complete words of up to n-grams. Consequently, it can better capture the complete semantics of the words constituting the sentence, thereby more accurately grasping the normal manifold information.
RoBERTa enhances BERT by implementing a dynamic masking strategy. This strategy involves duplicating the training data many times, resulting in each text being masked in many different ways throughout the epochs of pre-training. 
This allows for multiple masked copies of an example, enhancing the randomness of the input data and the learning capacity of the model. Therefore, RoBERTa can better focus on extracting the intrinsic knowledge about the manifold of normal examples.

\textbf{Pre-training techniques.} 
Compared to BERT and ALBERT, 
RoBERTa employs more sophisticated training techniques, such as longer training durations and larger batch sizes, enabling it to excel in completing the MLM task.
As such, the manifold of normal examples learned by RoBERTa aligns more closely with the underlying manifold of normal data, leading to improved performance of RoBERTa-based GradMLMD.

\subsection{ Contribution of Unmask Operation in Improving Detection Performance}
\label{subsec:important_unmask}


In this section, we assess the contribution of the unmask operation (or the masked language model) in detecting adversarial behaviors. Firstly, we remove the unmask operation directly from GradMLMD, creating GradMLMD-U. The results are presented in Table~\ref{tab:important_unmask}. Comparing GradMLMD and GradMLMD-U, it is evident that the removal of the unmask operation significantly degrades detection performance.
This outcome demonstrates that the masked language model plays a critical role in amplifying the difference between normal and adversarial examples.

\begin{table}[htbp]
  \centering
  \caption{The importance of the unmask operation in improving detection performance. 
  GradMLMD-U represents that only the masked texts 
  are used for extracting distinguishable scores. GRADMASK+U means adding the unmask operation (masked language models) to GRADMASK.
  }
  \resizebox{0.48\textwidth}{!}{
    \begin{tabular}{ccl|cc|cc|cc|cc}
    \toprule
    \multirow{2}[4]{*}{\textbf{Dataset}} & \multirow{2}[4]{*}{\textbf{Model}} & \multicolumn{1}{c}{\multirow{2}[4]{*}{\textbf{Method}}} & \multicolumn{2}{c}{\textbf{PWWS}} & \multicolumn{2}{c}{\textbf{TextFooler}} & \multicolumn{2}{c}{\textbf{TextBugger}} & \multicolumn{2}{c}{\textbf{DeepWordBug}} \\
\cmidrule{4-11}          &       & \multicolumn{1}{c}{} & \textbf{Acc.} & \multicolumn{1}{c}{\textbf{F1}} & \textbf{Acc.} & \multicolumn{1}{c}{\textbf{F1}} & \textbf{Acc.} & \multicolumn{1}{c}{\textbf{F1}} & \textbf{Acc.} & \textbf{F1} \\
    \midrule
    \multicolumn{1}{c}{\multirow{8}[4]{*}{\makecell{AG-\\NE\\WS}}} & \multicolumn{1}{c}{\multirow{4}[2]{*}{BERT}} & GradMLMD & 0.959  & 0.962  & \cellcolor[rgb]{ .988,  .894,  .839}0.979  & \cellcolor[rgb]{ .988,  .894,  .839}0.981  & \cellcolor[rgb]{ .988,  .894,  .839}0.951  & \cellcolor[rgb]{ .988,  .894,  .839}0.947  & \cellcolor[rgb]{ .988,  .894,  .839}0.940  & \cellcolor[rgb]{ .988,  .894,  .839}0.939  \\
          &       & GradMLMD-U & 0.903  & 0.905  & 0.947  & 0.950  & 0.901  & 0.905  & 0.878  & 0.889  \\
          &       & GRADMASK & 0.953  & 0.955  & 0.962  & 0.969  & 0.890  & 0.907  & 0.894  & 0.903  \\
          &       & GRADMASK+U & \cellcolor[rgb]{ .988,  .894,  .839}0.963  & \cellcolor[rgb]{ .988,  .894,  .839}0.970  & 0.967  & 0.972  & 0.910  & 0.916  & 0.919  & 0.917  \\
\cmidrule{2-11}          & \multicolumn{1}{c}{\multirow{4}[2]{*}{CNN}} & GradMLMD & \cellcolor[rgb]{ .988,  .894,  .839}0.961  & \cellcolor[rgb]{ .988,  .894,  .839}0.961  & \cellcolor[rgb]{ .988,  .894,  .839}0.970  & \cellcolor[rgb]{ .988,  .894,  .839}0.972  & \cellcolor[rgb]{ .988,  .894,  .839}0.950  & \cellcolor[rgb]{ .988,  .894,  .839}0.951  & \cellcolor[rgb]{ .988,  .894,  .839}0.951  & \cellcolor[rgb]{ .988,  .894,  .839}0.956  \\
          &       & GradMLMD-U & 0.930  & 0.933  & 0.940  & 0.946  & 0.921  & 0.920  & 0.902  & 0.909  \\
          &       & GRADMASK & 0.806  & 0.836  & 0.800  & 0.803  & 0.758  & 0.802  & 0.770  & 0.757  \\
          &       & GRADMASK+U & 0.854  & 0.842  & 0.872  & 0.873  & 0.833  & 0.842  & 0.856  & 0.845  \\
    \midrule
    \multicolumn{1}{c}{\multirow{8}[4]{*}{\makecell{IM\\DB}}} & \multicolumn{1}{c}{\multirow{4}[2]{*}{BERT}} & GradMLMD & 0.943  & 0.945  & \cellcolor[rgb]{ .988,  .894,  .839}0.950  & 0.951  & \cellcolor[rgb]{ .988,  .894,  .839}0.957  & \cellcolor[rgb]{ .988,  .894,  .839}0.955  & \cellcolor[rgb]{ .988,  .894,  .839}0.933  & \cellcolor[rgb]{ .988,  .894,  .839}0.935  \\
          &       & GradMLMD-U & 0.921  & 0.924  & 0.931  & 0.939  & 0.934  & 0.935  & 0.902  & 0.901  \\
          &       & GRADMASK & 0.943  & 0.943  & 0.922  & 0.925  & 0.907  & 0.911  & 0.870  & 0.875  \\
          &       & GRADMASK+U & \cellcolor[rgb]{ .988,  .894,  .839}0.949  & \cellcolor[rgb]{ .988,  .894,  .839}0.949  & 0.947  & \cellcolor[rgb]{ .988,  .894,  .839}0.953  & 0.930  & 0.932  & 0.911  & 0.912  \\
\cmidrule{2-11}          & \multicolumn{1}{c}{\multirow{4}[2]{*}{CNN}} & GradMLMD & \cellcolor[rgb]{ .988,  .894,  .839}0.906  & \cellcolor[rgb]{ .988,  .894,  .839}0.910  & \cellcolor[rgb]{ .988,  .894,  .839}0.921  & \cellcolor[rgb]{ .988,  .894,  .839}0.919  & \cellcolor[rgb]{ .988,  .894,  .839}0.913  & \cellcolor[rgb]{ .988,  .894,  .839}0.921  & \cellcolor[rgb]{ .988,  .894,  .839}0.899  & \cellcolor[rgb]{ .988,  .894,  .839}0.903  \\
          &       & GradMLMD-U & 0.862  & 0.864  & 0.892  & 0.896  & 0.869  & 0.886  & 0.822  & 0.828  \\
          &       & GRADMASK & 0.794  & 0.789  & 0.750  & 0.755  & 0.721  & 0.766  & 0.706  & 0.732  \\
          &       & GRADMASK+U & 0.844  & 0.832  & 0.801  & 0.821  & 0.793  & 0.802  & 0.765  & 0.784  \\
    \bottomrule
    \end{tabular}%
    }
  \label{tab:important_unmask}%
\end{table}%

We also compare GradMLMD and GradMLMD-U with GRADMASK~\cite{Moon2022GradMaskGT},
which distinguishes from GradMLMD in two aspects: (1) GRADMASK excludes the unmask operation, and (2) it utilizes a masking strategy that concurrently masks multiple words in the input during a mask operation.

We first compare GradMLMD-U with GRADMASK. Both methods do not use masked language models, and they primarily differ in their masking strategies. 
The results of table~\ref{tab:important_unmask} show that GradMLMD-U outperforms GRADMASK by $6.3\%$ (Acc.) and $5.6\%$ (F1) in average. 
As masking multiple words simultaneously introduces significant noise into the input, it reduces the maliciousness of adversarial inputs and compromises the naturalness of normal texts.

We then consider incorporating the unmask operation (masked language models) into the existing detector GRADMASK, resulting in GRADMASK+U. The results consistently demonstrate improvements in detection scores over the original GRADMASK, aligning with the earlier comparison between GradMLMD and GradMLMD-U.

GradMLMD also generally outperforms GRADMASK+U in most cases due to variations in the masked sequences resulting from their different masking strategies. 
For example,  GradMLMD produces a masked sequence $\{w_{\textnormal{1}}, w_{\textnormal{2}}, \cdots,[MASK], \cdots, w_{n}\}$, while GRADMASK+U produces $\{[MASK], [MASK], \cdots,[MASK], \cdots, w_{n}\}$. 
In the former one, the masked language model is tasked with predicting a single word, utilizing its learned manifold knowledge. In contrast, the latter one requires the model to simultaneously fill in multiple missing pieces of information, posing a more significant challenge. Consequently, this approach may struggle to correctly project the masked example back to the manifold where normal data is located. 

\subsection{Impact of the Number of Unmask Operation Reconstruction (\texorpdfstring{$k$}{})}
\label{subsec:setting_unmask}

As highlighted in Sec.~\ref{Sec:mlmd},
during detection, MLMD requires selecting the top-$k$ candidates from unmasking results, each triggering a single invocation of the victim model. To minimize resource usage, the smallest viable $k$ maintaining detection performance is crucial.

The impact of the number of reconstruction ($k$) of the masked language model on the detection results is illustrated in Fig.~\ref{Fig:gradmlmd_differences}(d-f). The result indicates that $k=3$ is the most favorable setting for detecting performance. Meanwhile, $k=1$ is also a reasonable choice, as there is a slight decrease in the detection scores when reconstructing only once. However, it still significantly outperforms comparison algorithms. Therefore, simply masking keywords and reconstructing these words only once is a practical solution for applications requiring stricter response time.

Fig.~\ref{Fig:gradmlmd_differences}(d-f), the detection F1 score decreases when $k>3$. This decline may stem from the weakened ability of $k$ reconstruction examples to induce manifold changes. Examples reconstructed with words receiving lower confidence scores from masked language models may not facilitate useful manifold transformations for detection. Instead, they introduce noise that complicates the detection process.

\subsection{Fine-tuning Masked Language Models for Detection}
\label{subsec:fine_tuning}
\begin{table}[htbp]
  \centering
  \caption{Detection results of the fine-tuned masked language model as the component of GradMLMD. GradMLMD-F denotes GradMLMD instantiated by the fine-tuned RoBERTa.}
  \resizebox{0.48\textwidth}{!}{
    \begin{tabular}{ccl|cc|cc|cc|cc}
    \toprule
    \multirow{2}[4]{*}{\textbf{Dataset}} & \multirow{2}[4]{*}{\textbf{Model}} & \multicolumn{1}{c}{\multirow{2}[4]{*}{\textbf{Method}}} & \multicolumn{2}{c}{\textbf{PWWS}} & \multicolumn{2}{c}{\textbf{TextFooler}} & \multicolumn{2}{c}{\textbf{TextBugger}} & \multicolumn{2}{c}{\textbf{DeepWordBug}} \\
\cmidrule{4-11}          &       & \multicolumn{1}{c}{} & \textbf{Acc.} & \multicolumn{1}{c}{\textbf{F1}} & \textbf{Acc.} & \multicolumn{1}{c}{\textbf{F1}} & \textbf{Acc.} & \multicolumn{1}{c}{\textbf{F1}} & \textbf{Acc.} & \textbf{F1} \\
    \midrule
    \multirow{8}[8]{*}{\makecell{AG-\\NE\\WS}} & \multirow{2}[2]{*}{BERT} & GradMLMD & 0.959  & 0.962  & \cellcolor[rgb]{ .988,  .894,  .839}0.979  & \cellcolor[rgb]{ .988,  .894,  .839}0.981  & 0.951  & 0.947  & 0.940  & 0.939  \\
          &       & GradMLMD-F & \cellcolor[rgb]{ .988,  .894,  .839}0.962  & \cellcolor[rgb]{ .988,  .894,  .839}0.965  & 0.977  & 0.978  & \cellcolor[rgb]{ .988,  .894,  .839}0.957  & \cellcolor[rgb]{ .988,  .894,  .839}0.955  & \cellcolor[rgb]{ .988,  .894,  .839}0.945  & \cellcolor[rgb]{ .988,  .894,  .839}0.942  \\
\cmidrule{2-11}          & \multirow{2}[2]{*}{ALBERT} & GradMLMD & 0.951  & 0.952  & 0.985  & 0.987  & 0.961  & 0.963  & 0.940  & 0.942  \\
          &       & GradMLMD-F & \cellcolor[rgb]{ .988,  .894,  .839}0.969  & \cellcolor[rgb]{ .988,  .894,  .839}0.971  & \cellcolor[rgb]{ .988,  .894,  .839}0.985  & \cellcolor[rgb]{ .988,  .894,  .839}0.987  & \cellcolor[rgb]{ .988,  .894,  .839}0.969  & \cellcolor[rgb]{ .988,  .894,  .839}0.971  & \cellcolor[rgb]{ .988,  .894,  .839}0.952  & \cellcolor[rgb]{ .988,  .894,  .839}0.956  \\
\cmidrule{2-11}          & \multirow{2}[2]{*}{CNN} & GradMLMD & 0.961  & 0.961  & 0.970  & 0.972  & 0.950  & 0.951  & 0.951  & 0.956  \\
          &       & GradMLMD-F & \cellcolor[rgb]{ .988,  .894,  .839}0.966  & \cellcolor[rgb]{ .988,  .894,  .839}0.967  & \cellcolor[rgb]{ .988,  .894,  .839}0.986  & \cellcolor[rgb]{ .988,  .894,  .839}0.987  & \cellcolor[rgb]{ .988,  .894,  .839}0.965  & \cellcolor[rgb]{ .988,  .894,  .839}0.961  & \cellcolor[rgb]{ .988,  .894,  .839}0.961  & \cellcolor[rgb]{ .988,  .894,  .839}0.972  \\
\cmidrule{2-11}          & \multirow{2}[2]{*}{LSTM} & GradMLMD & 0.942  & 0.944  & 0.960  & 0.962  & 0.949  & 0.949  & 0.943  & 0.941  \\
          &       & GradMLMD-F & \cellcolor[rgb]{ .988,  .894,  .839}0.961  & \cellcolor[rgb]{ .988,  .894,  .839}0.963  & \cellcolor[rgb]{ .988,  .894,  .839}0.967  & \cellcolor[rgb]{ .988,  .894,  .839}0.971  & \cellcolor[rgb]{ .988,  .894,  .839}0.969  & \cellcolor[rgb]{ .988,  .894,  .839}0.969  & \cellcolor[rgb]{ .988,  .894,  .839}0.960  & \cellcolor[rgb]{ .988,  .894,  .839}0.959  \\
    \midrule
    \multirow{8}[8]{*}{\makecell{IM\\DB}} & \multirow{2}[2]{*}{BERT} & GradMLMD & 0.943  & 0.945  & 0.950  & 0.951  & 0.957  & 0.955  & 0.933  & 0.935  \\
          &       & GradMLMD-F & \cellcolor[rgb]{ .988,  .894,  .839}0.959  & \cellcolor[rgb]{ .988,  .894,  .839}0.970  & \cellcolor[rgb]{ .988,  .894,  .839}0.956  & \cellcolor[rgb]{ .988,  .894,  .839}0.957  & \cellcolor[rgb]{ .988,  .894,  .839}0.958  & \cellcolor[rgb]{ .988,  .894,  .839}0.959  & \cellcolor[rgb]{ .988,  .894,  .839}0.946  & \cellcolor[rgb]{ .988,  .894,  .839}0.949  \\
\cmidrule{2-11}          & \multirow{2}[2]{*}{ALBERT} & GradMLMD & \cellcolor[rgb]{ .988,  .894,  .839}0.947  & \cellcolor[rgb]{ .988,  .894,  .839}0.950  & \cellcolor[rgb]{ .988,  .894,  .839}0.970  & \cellcolor[rgb]{ .988,  .894,  .839}0.972  & 0.951  & 0.953  & 0.893  & 0.895  \\
          &       & GradMLMD-F & 0.945  & 0.950  & 0.970  & 0.970  & \cellcolor[rgb]{ .988,  .894,  .839}0.962  & \cellcolor[rgb]{ .988,  .894,  .839}0.961  & \cellcolor[rgb]{ .988,  .894,  .839}0.923  & \cellcolor[rgb]{ .988,  .894,  .839}0.925  \\
\cmidrule{2-11}          & \multirow{2}[2]{*}{CNN} & GradMLMD & \cellcolor[rgb]{ .988,  .894,  .839}0.906  & \cellcolor[rgb]{ .988,  .894,  .839}0.910  & 0.921  & 0.919  & 0.913  & 0.921  & 0.899  & 0.903  \\
          &       & GradMLMD-F & 0.898  & 0.903  & \cellcolor[rgb]{ .988,  .894,  .839}0.931  & \cellcolor[rgb]{ .988,  .894,  .839}0.932  & \cellcolor[rgb]{ .988,  .894,  .839}0.933  & \cellcolor[rgb]{ .988,  .894,  .839}0.936  & \cellcolor[rgb]{ .988,  .894,  .839}0.904  & \cellcolor[rgb]{ .988,  .894,  .839}0.909  \\
\cmidrule{2-11}          & \multirow{2}[2]{*}{LSTM} & GradMLMD & 0.882  & 0.884  & 0.906  & 0.918  & 0.903  & 0.904  & 0.886  & 0.897  \\
          &       & GradMLMD-F & \cellcolor[rgb]{ .988,  .894,  .839}0.917  & \cellcolor[rgb]{ .988,  .894,  .839}0.914  & \cellcolor[rgb]{ .988,  .894,  .839}0.914  & \cellcolor[rgb]{ .988,  .894,  .839}0.926  & \cellcolor[rgb]{ .988,  .894,  .839}0.930  & \cellcolor[rgb]{ .988,  .894,  .839}0.934  & \cellcolor[rgb]{ .988,  .894,  .839}0.896  & \cellcolor[rgb]{ .988,  .894,  .839}0.907  \\
    \bottomrule
    \end{tabular}%
    }
  \label{tab:de_fine}%
\end{table}%

Prior works \cite{Gururangan2020DontSP} report that performing the MLM objective on the target domain with unlabeled data can also help to improve downstream task performance. 
We therefore explore the possibility of fine-tuning the masked language model $\Phi$ to further enhance the detection performance of GradMLMD. With all other parameters kept constant, we fine-tune RoBERTa on the target domains (AG-NEWS, IMDB), and the detector based on these models is referred to as GradMLMD-F. 
Due to restricted computing resources, we train RoBERTa for $20$ epochs instead of the original $40$ and decrease the batch size from $8000$ to $128$.

The experimental results present in Table \ref{tab:de_fine} demonstrate a clear upward trend in performance for GradMLMD-F when compared to the GradMLMD, with an average increase of $1.4\%$ and $1.5\%$ in terms of accuracy and F1 score, respectively. After fine-tuning, the manifold fitted by RoBERTa is better aligned with the manifold of the target domain, resulting in improved detection ability.
This finding indicates that further performance enhancements hinge on techniques that can equip the masked language model with more precise manifold knowledge of the target domain.

\section{Discussion} 

If attackers realize the existence of defense methods, they can adjust their strategies to build adaptive attacks that
optimize adversarial perturbations to maximize victim model loss while evading MLMD/GradMLMD detectors. This task is akin to generating adversarial examples whose manifold remains unchanged after mask and unmask operations. Nevertheless, the masked language model only models the manifold of normal texts, conflicting with the attacker's goal. Consequently, depending on how well the masked language model fits the manifold of normal data, our detectors can substantially raise the cost of adaptive attacks or potentially render them ineffective.

\section{Conclusion}
This work first introduces an initial textual adversarial example detector, MLMD, drawing from the insight that masked language models can capture the manifold of normal examples and the off-manifold nature of adversarial examples. 
While MLMD shows superior detection, its deployment is challenging. 
We then reveal the existence and influence of non-keywords for detection performance,
exploit gradient signals to locate and filter out non-keywords practically and introduce Gradient-guided MLMD (GradMLMD), which specifically applies manifold changes to keywords only. Extensive experiments  
demonstrate that GradMLMD consistently exhibits similar detection capabilities in line with those of MLMD, while significantly reducing resource overhead.

The recent achievements of large language models (LLMs) have brought the field of natural language understanding into a new era. 
Our study represents an initial effort to employ masked language models for detecting adversarial inputs, yielding promising results. The effective utilization of advanced LLMs to bolster adversarial robustness will be the focus of future research.

\footnotesize{
\bibliographystyle{IEEEtran}
\bibliography{taslp_arxiv}}
\vspace{-1.5cm}

\clearpage

\small{
\section*{Supplementary Document}
\label{sec:supp_doc}



\subsection{Datasets}
\label{subsec:datasets_doc}
\textbf{AG-NEWS:} 
The dataset comprises news articles categorized into four topics: World, Sports, Business, and Sci/Tech. It consists of 120K articles in the training set and 7.6K in the test set, with an average article length of 43 words.

\textbf{IMDB:} 
The dataset contains 50K movie reviews aimed at binary sentiment classification (positive or negative). It is split into a training set of 25K reviews and a test set of 25K reviews. The average review length in IMDB is 215 words.

\textbf{SST-2:} 
SST, a corpus with fully labeled parse trees, is tailored for analyzing the compositional effects of sentiment.
Following the setting in \cite{Moon2022GradMaskGT, Mozes2021FrequencyGuidedWS}, we transform it into a binary dataset (SST-2) annotated with positive or negative labels. 
The dataset comprises a training set with 67K texts, a validation set with 0.8K sequences, and a test set with 1.8K texts. On average, a text spans 20 words in length.

\subsection{Attack Methods}
\label{subsec:attack_doc}
\textbf{PWWS} (word-level): Probability weighted word saliency \cite{Ren2019GeneratingNL} is a greedy algorithm based on a synonym replacement strategy that introduces a novel word replacement order determined by both the word saliency and the classification probability. 

\textbf{TextFooler} (word-level): TextFooler \cite{Jin2020IsBR} 
initially ranks words in the input text according to their importance. Subsequently, it substitutes these words with semantically similar and grammatically appropriate alternatives until a change in prediction occurs.

\textbf{TextBugger} (character-level): TextBugger \cite{Li2019TextBuggerGA} considers a more general framework of deep learning-based text understanding. It identifies the pivotal token for manipulation and then chooses the most suitable perturbation from five alternatives.

\textbf{DeepWordBug} (character-level): DeepWordBug \cite{Gao2018BlackBoxGO} introduces a novel scoring function to identify crucial words.
Subsequently, simple character-level transformations are applied to the top-ranked words to minimize the perturbation's edit distance.

        

\subsection{Compared Detectors}
\label{subsec:detector_doc}
\textbf{FGWS:} Based on the assumption that adversarial attacks prefer to replace words in the input text with low-frequency words from the training set to induce adversarial behaviors, FGWS \cite{Mozes2021FrequencyGuidedWS} replaces words whose frequency is below a pre-defined threshold with higher-frequency synonyms. By nature, FGWS is only suitable for word-level attacks. 

\textbf{WDR:} Inspired by the use of logits-based adversarial detectors in computer vision tasks, WDR \cite{Mosca2022ThatIA} quantifies the impact of words via the word-level differential reaction (in logits) and then trains an adversarial classifier over a reaction dataset generated from both normal and adversarial examples. Unlike FGWS, WDR is applicable for detecting both word-level and character-level attacks. 

\textbf{GRADMASK:} GRADMASK \cite{Moon2022GradMaskGT} identifies some important tokens using gradient signals and subsequently occludes them with the $[MASK]$ token. The masked sequences are then fed into the victim model to assess the change in the model's confidence regarding the prediction of the original input. 
GRADMASK is also capable of detecting attacks at both the character-level and word-level. 
The main differences between GRADMASK and our MLMD/GradMLMD lie in the masking strategy employed in the mask operation and the use of a masked language model. 

We adjust their original implementations and fine-tune parameters to obtain the best results.



\subsection{The Detection Performance of the Threshold-based Classifier for MLMD}
Table \ref{tab:de_pre_sst2} displays the detection results of FGWS, WDR, GRADMASK, and MLMD on SST-2 dataset. Consistent with the findings from the AG-NEWS and IMDB, our approach shows superior accuracy and F1 score compared to FGWS, WDR, and GRADMASK in most scenarios. This demonstrates the advantage of using masked language models for detection. We do not report the detection scores of FGWS under TextBugger and DeepWordBug attacks. Because it is challenging for FGWS to find suitable high-frequency synonyms for replacement, resulting in less reliable detection outcomes.
\begin{table*}[htbp]
  \centering
  \caption{Detection performance of FGWS, WDR, GRADMASK, and MLMD on SST-2. We omit the detection results of FGWS for adversarial examples generated by TextBugger attack and DeepWordBug attack, as it struggles to locate appropriate synonyms from training sets for certain words when only characters are perturbed.}
  \resizebox{0.76\textwidth}{!}{
    \begin{tabular}{ccl|cc|cc|cc|cc}
    \toprule
    \multirow{2}[4]{*}{\textbf{Dataset}} & \multirow{2}[4]{*}{\textbf{Model}} & \multicolumn{1}{l}{\multirow{2}[4]{*}{\textbf{Method}}} & \multicolumn{2}{c}{\textbf{PWWS}} & \multicolumn{2}{c}{\textbf{TextFooler}} & \multicolumn{2}{c}{\textbf{TextBugger}} & \multicolumn{2}{c}{\textbf{DeepWordBug}} \\
\cmidrule{4-11}          &       & \multicolumn{1}{c}{} & \textbf{Acc.} & \multicolumn{1}{c}{\textbf{F1}} & \textbf{Acc.} & \multicolumn{1}{c}{\textbf{F1}} & \textbf{Acc.} & \multicolumn{1}{c}{\textbf{F1}} & \textbf{Acc.} & \textbf{F1} \\
    \midrule
    \multicolumn{1}{c}{\multirow{16}[8]{*}{SST-2}} & \multirow{4}[2]{*}{BERT} & FGWS  & 0.821  & 0.790  & 0.769  & 0.711  &        -    &        -    &        -    &        -    \\
          &       & WDR   & 0.787  & 0.800  & 0.790  & 0.807  & 0.783  & 0.807  & 0.728  & 0.737  \\
          &       & GRADMASK & 0.765  & 0.792  & 0.801  & 0.816  & 0.790  & 0.804  & 0.750  & 0.776  \\
          &       & MLMD  & \cellcolor[rgb]{ .988,  .894,  .839}0.837  & \cellcolor[rgb]{ .988,  .894,  .839}0.848  & \cellcolor[rgb]{ .988,  .894,  .839}0.874  & \cellcolor[rgb]{ .988,  .894,  .839}0.879  & \cellcolor[rgb]{ .988,  .894,  .839}0.852  & \cellcolor[rgb]{ .988,  .894,  .839}0.855  & \cellcolor[rgb]{ .988,  .894,  .839}0.830  & \cellcolor[rgb]{ .988,  .894,  .839}0.839  \\
\cmidrule{2-11}          & \multirow{4}[2]{*}{ALBERT} & FGWS  & 0.811  & 0.779  & 0.738  & 0.663  &        -    &        -    &        -    &        -    \\
          &       & WDR   & 0.730  & 0.757  & 0.747  & 0.773  & 0.714  & 0.755  & 0.674  & 0.698  \\
          &       & GRADMASK & 0.768  & 0.787  & 0.751  & 0.805  & 0.761  & 0.785  & 0.739  & 0.803  \\
          &       & MLMD  & \cellcolor[rgb]{ .988,  .894,  .839}0.837  & \cellcolor[rgb]{ .988,  .894,  .839}0.848  & \cellcolor[rgb]{ .988,  .894,  .839}0.850  & \cellcolor[rgb]{ .988,  .894,  .839}0.857  & \cellcolor[rgb]{ .988,  .894,  .839}0.849  & \cellcolor[rgb]{ .988,  .894,  .839}0.855  & \cellcolor[rgb]{ .988,  .894,  .839}0.826  & \cellcolor[rgb]{ .988,  .894,  .839}0.836  \\
\cmidrule{2-11}          & \multirow{4}[2]{*}{CNN} & FGWS  & 0.798  & 0.767  & 0.689  & 0.588  &        -    &        -    &        -    &        -    \\
          &       & WDR   & 0.715  & 0.769  & 0.719  & 0.773  & 0.703  & 0.768  & 0.721  & 0.776  \\
          &       & GRADMASK & 0.720  & 0.727  & 0.726  & 0.729  & 0.711  & 0.716  & 0.713  & 0.715  \\
          &       & MLMD  & \cellcolor[rgb]{ .988,  .894,  .839}0.811  & \cellcolor[rgb]{ .988,  .894,  .839}0.829  & \cellcolor[rgb]{ .988,  .894,  .839}0.851  & \cellcolor[rgb]{ .988,  .894,  .839}0.859  & \cellcolor[rgb]{ .988,  .894,  .839}0.842  & \cellcolor[rgb]{ .988,  .894,  .839}0.850  & \cellcolor[rgb]{ .988,  .894,  .839}0.817  & \cellcolor[rgb]{ .988,  .894,  .839}0.831  \\
\cmidrule{2-11}          & \multirow{4}[2]{*}{LSTM} & FGWS  & \cellcolor[rgb]{ .988,  .894,  .839}0.801  & \cellcolor[rgb]{ .988,  .894,  .839}0.811  & 0.678  & 0.566  &        -    &        -    &        -    &        -    \\
          &       & WDR   & 0.736  & 0.776  & 0.751  & 0.788  & 0.724  & 0.778  & 0.686  & 0.742  \\
          &       & GRADMASK & 0.739  & 0.742  & 0.713  & 0.736  & 0.789  & 0.790  & 0.705  & 0.718  \\
          &       & MLMD  & 0.778  & 0.809  & \cellcolor[rgb]{ .988,  .894,  .839}0.832  & \cellcolor[rgb]{ .988,  .894,  .839}0.843  & \cellcolor[rgb]{ .988,  .894,  .839}0.832  & \cellcolor[rgb]{ .988,  .894,  .839}0.843  & \cellcolor[rgb]{ .988,  .894,  .839}0.788  & \cellcolor[rgb]{ .988,  .894,  .839}0.805  \\
    \bottomrule
    \end{tabular}%
    }
  \label{tab:de_pre_sst2}%
\end{table*}%


\subsection{The Detection Performance of the Model-based Classifier for MLMD}

The detection performance of various model-based classifiers on SST-2, AG-NEWS, and IMDB is presented in Table \ref{tab:model_pre}. It's clear that both model-based classifiers and threshold-based classifiers exhibit comparable detection capabilities. Furthermore, as anticipated, classifiers trained on $\overline{\Gamma}$ showcase superior effectiveness.
This suggests that sorting feature vectors facilitates classification compared to using the original vectors, as the former includes extra information regarding the arrangement of vector elements.

\begin{table*}[htbp]
  \centering
  \caption{The detection performance of MLMD was constructed using model-based classifiers on SST-2, AG-NEWS and IMDB. Model-C and Model columns indicate the architecture of the adversarial classifier and the victim model, respectively. Type denotes the different datasets ($\Gamma$ and $\overline{\Gamma}$ in Sec.~\ref{subsubsec:ModelbasedDetector}) used for training the adversarial classifier.}
   \resizebox{0.86\textwidth}{!}{
    \begin{tabular}{cccccccccccc}
    \toprule
    \multicolumn{1}{c}{\multirow{2}[4]{*}{\textbf{Model-C}}} & \multicolumn{1}{c}{\multirow{2}[4]{*}{\textbf{Dataset}}} & \multicolumn{1}{c}{\multirow{2}[4]{*}{\textbf{Model}}} & \multirow{2}[4]{*}{\textbf{Type}} & \multicolumn{2}{c}{\textbf{PWWS}} & \multicolumn{2}{c}{\textbf{TextFooler}} & \multicolumn{2}{c}{\textbf{TextBugger}} & \multicolumn{2}{c}{\textbf{DeepWordBug}} \\
\cmidrule{5-12}          &       &       & \multicolumn{1}{c}{} & \textbf{Acc.} & \textbf{F1.} & \textbf{Acc.} & \textbf{F1.} & \textbf{Acc.} & \textbf{F1.} & \textbf{Acc.} & \textbf{F1.} \\
    \midrule
    \multicolumn{1}{c}{\multirow{12}[6]{*}{MLP}} & \multicolumn{1}{c}{\multirow{4}[2]{*}{SST-2}} & \multicolumn{1}{c}{\multirow{2}[1]{*}{BERT}} & \textbf{$\Gamma$}  & 0.813  & 0.820  & 0.824  & 0.815  & 0.846  & 0.837  & 0.827  & 0.819  \\
          &       &       & \textbf{$\overline{\Gamma}$}     & 0.831  & 0.824  & 0.844  & 0.842  & 0.860  & 0.855  & 0.815  & 0.801  \\
          &       & \multicolumn{1}{c}{\multirow{2}[1]{*}{CNN}} & \textbf{$\Gamma$}    & 0.807  & 0.814  & 0.829  & 0.824  & 0.808  & 0.799  & 0.789  & 0.799  \\
          &       &       & \textbf{$\overline{\Gamma}$}     & 0.809  & 0.811  & 0.854  & 0.861  & 0.844  & 0.848  & 0.801  & 0.789  \\
\cmidrule{2-12}          & \multicolumn{1}{c}{\multirow{4}[2]{*}{AG-NEWS}} & \multicolumn{1}{c}{\multirow{2}[1]{*}{BERT}} & \textbf{$\Gamma$}     & 0.942  & 0.937  & 0.960  & 0.955  & 0.946  & 0.940  & 0.930  & 0.930  \\
          &       &       & \textbf{$\overline{\Gamma}$}     & 0.958  & 0.956  & 0.974  & 0.972  & 0.938  & 0.933  & 0.942  & 0.944  \\
          &       & \multicolumn{1}{c}{\multirow{2}[1]{*}{CNN}} & \textbf{$\Gamma$}     & 0.970  & 0.969  & 0.966  & 0.965  & 0.932  & 0.931  & 0.924  & 0.921  \\
          &       &       & \textbf{$\overline{\Gamma}$}     & 0.961  & 0.960  & 0.965  & 0.965  & 0.959  & 0.960  & 0.937  & 0.935  \\
\cmidrule{2-12}          & \multicolumn{1}{c}{\multirow{4}[2]{*}{IMDB}} & \multicolumn{1}{c}{\multirow{2}[1]{*}{BERT}} & \textbf{$\Gamma$}     & 0.883  & 0.866  & 0.864  & 0.837  & 0.916  & 0.902  & 0.845  & 0.823  \\
          &       &       & \textbf{$\overline{\Gamma}$}     & 0.946  & 0.946  & 0.937  & 0.935  & 0.941  & 0.938  & 0.934  & 0.934  \\
          &       & \multicolumn{1}{c}{\multirow{2}[1]{*}{CNN}} & \textbf{$\Gamma$}     & 0.853  & 0.851  & 0.922  & 0.925  & 0.923  & 0.924  & 0.893  & 0.893  \\
          &       &       & \textbf{$\overline{\Gamma}$}     & 0.881  & 0.885  & 0.930  & 0.933  & 0.927  & 0.929  & 0.922  & 0.923  \\
    \midrule
    \multicolumn{1}{c}{\multirow{12}[6]{*}{XGBoost}} & \multicolumn{1}{c}{\multirow{4}[2]{*}{SST-2}} & \multicolumn{1}{c}{\multirow{2}[1]{*}{BERT}} & \textbf{$\Gamma$}     & 0.831  & 0.829  & 0.886  & 0.882  & 0.877  & 0.897  & 0.853  & 0.845  \\
          &       &       &\textbf{$\overline{\Gamma}$}     & 0.837  & 0.833  & 0.889  & 0.883  & 0.887  & 0.889  & 0.863  & 0.866  \\
          &       & \multicolumn{1}{c}{\multirow{2}[1]{*}{CNN}} & \textbf{$\Gamma$}     & 0.827  & 0.829  & 0.854  & 0.866  & 0.843  & 0.853  & 0.827  & 0.828  \\
          &       &       & \textbf{$\overline{\Gamma}$}    & 0.859  & 0.860  & 0.873  & 0.884  & 0.859  & 0.866  & 0.841  & 0.843  \\
\cmidrule{2-12}          & \multicolumn{1}{c}{\multirow{4}[2]{*}{AG-NEWS}} & \multicolumn{1}{c}{\multirow{2}[1]{*}{BERT}} &  \textbf{$\Gamma$}    & 0.944  & 0.940  & 0.956  & 0.953  & 0.935  & 0.939  & 0.937  & 0.948  \\
          &       &       & \textbf{$\overline{\Gamma}$}   & 0.956  & 0.964  & 0.973  & 0.971  & 0.948  & 0.944  & 0.958  & 0.960  \\
          &       & \multicolumn{1}{c}{\multirow{2}[1]{*}{CNN}} & \textbf{$\Gamma$}     & 0.960  & 0.959  & 0.963  & 0.963  & 0.950  & 0.947  & 0.933  & 0.932  \\
          &       &       & \textbf{$\overline{\Gamma}$}    & 0.978  & 0.978  & 0.978  & 0.976  & 0.967  & 0.967  & 0.951  & 0.951  \\
\cmidrule{2-12}          & \multicolumn{1}{c}{\multirow{4}[2]{*}{IMDB}} & \multicolumn{1}{c}{\multirow{2}[1]{*}{BERT}} &  \textbf{$\Gamma$}     & 0.929  & 0.929  & 0.934  & 0.938  & 0.937  & 0.935  & 0.927  & 0.927  \\
          &       &       & \textbf{$\overline{\Gamma}$}     & 0.963  & 0.953  & 0.946  & 0.944  & 0.955  & 0.954  & 0.948  & 0.948  \\
          &       & \multicolumn{1}{c}{\multirow{2}[1]{*}{CNN}} & \textbf{$\Gamma$}     & 0.857  & 0.867  & 0.935  & 0.939  & 0.931  & 0.935  & 0.898  & 0.901  \\
          &       &       & \textbf{$\overline{\Gamma}$}     & 0.935  & 0.938  & 0.952  & 0.955  & 0.955  & 0.957  & 0.938  & 0.938  \\
    \bottomrule
    \end{tabular}%
    }
  \label{tab:model_pre}%
\end{table*}%
}

\end{document}